\documentclass{article}

     \PassOptionsToPackage{numbers, compress}{natbib}

\usepackage[eandd, final]{neurips_2026}

\usepackage[utf8]{inputenc} 
\usepackage[T1]{fontenc}    
\usepackage{hyperref}       
\usepackage{url}            
\usepackage{booktabs}       
\usepackage{amsfonts}       
\usepackage{nicefrac}       
\usepackage{microtype}      
\usepackage{xcolor}         
\usepackage{graphicx}
\usepackage{amsmath}
\usepackage{wrapfig}    
\usepackage{todonotes}
\usepackage{wasysym}
\usepackage{multirow}
\usepackage{threeparttable}
\usepackage{colortbl}
\usepackage{xcolor}
\usepackage{enumitem}

\title{GVGAI-LLM: Evaluating Large Language Models with Infinite Games}

\author{%
  Yuchen Li\thanks{\texttt{yl6394@nyu.edu}. This work is currently under review at NeurIPS 2026. Available at: \url{https://github.com/doveliyuchen/GVGAI_GYM}} \\
  New York University \\
  \And
  Cong Lin \\
  Independent Researcher \\
  \And
  Muhammad Umair Nasir \\
  New York University \\
  University of the Witwatersrand \\
  \And
  Philip Bontrager \\
  Meta \\
  \And
  Jialin Liu \\
  Lingnan University \\
  \And
  Julian Togelius \\
  New York University \\
}

\begin{document}

\maketitle

\begin{abstract}
  We introduce GVGAI-LLM, a video game benchmark for evaluating the reasoning and problem-solving capabilities of large language models (LLMs). Built on the General Video Game AI framework, it features a diverse collection of arcade-style games designed to test a model’s ability to handle tasks that differ from most existing LLM benchmarks. The benchmark leverages a video game description language that enables the rapid creation of new games (including rules and levels, helping to prevent overfitting over time. Each game scene is represented by a compact set of ASCII characters, allowing for efficient processing by language models. GVGAI-LLM defines interpretable metrics, including meaningful step ratio, step efficiency, and overall score, to assess model behavior. Through zero-shot evaluations across 118 games with diverse challenges and skill depth, we reveal persistent limitations of LLMs in spatial reasoning and basic planning. Current models consistently exhibit spatial and logical errors, motivating structured prompting and spatial grounding techniques. Although these interventions lead to partial improvements, the benchmark remains very far from being solved. GVGAI-LLM serves a reproducible testbed for advancing research on language model capabilities, with a particular emphasis on agentic behavior and spatial reasoning. Furthermore, its ability to generate infinite benchmarks, both manually and procedurally, provides a scalable framework for longitudinal evaluation.
\end{abstract}

\section{Introduction}

Large language models (LLMs) have shown impressive capabilities in open-ended generation, planning, and reasoning tasks~\cite{achiam2023gpt,liu2025deepseek,singh2025openai,guan2025monitoring}. LLMs are also deployed as interactive agents in sequential decision making environments such as robotics, web navigation, and games~\cite{achiam2023gpt,10974629,topsakal2024evaluating}. While benchmarks such as GLUE~\cite{wang2018glue}, SuperGLUE~\cite{NEURIPS2019_4496bf24}, and MMLU~\cite{hendrycks2021measuring} evaluate static language understanding and broad domain knowledge, HumanEval~\cite{chen2021evaluatinglargelanguagemodels} assesses code generation proficiency, HELM~\cite{liang2023holistic} examines safety and robustness, and instruction-following frameworks like ALFWorld~\cite{shridhar2021alfworld} and BabyAI~\cite{chevalier-boisvert2018babyai} focus on executing textual commands. However, many reasoning and knowledge benchmarks show signs of saturation. Also, these benchmarks measure decision making competence in structured symbolic domains with game-style logic, real-time decision, and spatial reasoning. 

\begin{figure*}[htbp]
  \centering
    \includegraphics[width=0.8\linewidth,trim= 1cm 6.0cm 1cm 6.0cm, clip]{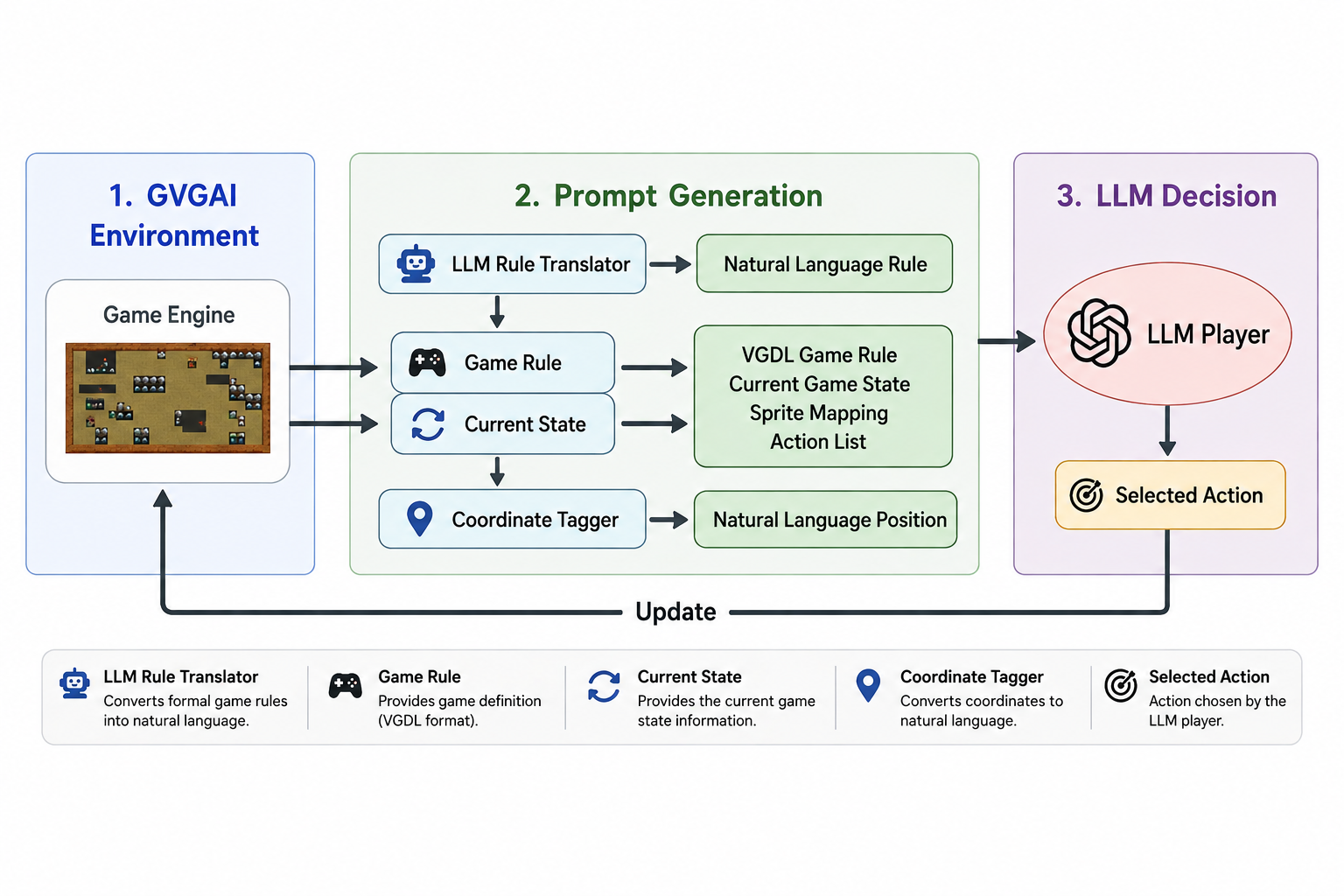}
  \caption{Overview of GVGAI-LLM}
  \label{fig:overall}
\end{figure*}

To address this gap, we introduce \textbf{GVGAI-LLM} (c.f., Figure \ref{fig:overall}), a benchmark that adapts the General Video Game AI (GVGAI) framework~\cite{perez2019general,liebana2019general2} into a testbed tailored for LLMs. Unlike existing benchmarks focused on task completion or instruction following, GVGAI-LLM emphasizes goal-directed agent behavior in reactive, rule-based 2D games. The GVGAI framework’s diverse game dynamics and formal Video Game Description Language (VGDL) specifications~\cite{ebner2013towards,schaul2013video} make it ideal for evaluating how well LLMs can parse game environments, comply with goals, and act.
Our key contributions are as follows:
\begin{itemize}
\item We introduce GVGAI-LLM, a standardized system that enables large language model-based agents to interact with more than one hundred games via natural language, supporting zero-shot and contextual prompting. 
Symbolic game states are encoded into structured textual representations, making them accessible to language-only agents without reliance on internal simulators or programmatic logic.
\item We design interpretable and reproducible metrics, including meaningful step ratio, step efficiency, and win rate, to assess agent behavior and reveal key failure modes.
\item Extensive benchmarking of nine LLMs against search-based and reinforcement learning (RL) agents across diverse games reveals persistent deficiencies in spatial reasoning and basic planning. These findings position GVGAI-LLM as a robust, reproducible testbed to advance research into behavior and the fundamental cognitive limits of language models.
\end{itemize}
GVGAI-LLM\footnote{The code is available through anonymous GitHub in supplementary material.} provides a unified testbed that supports natural language interfaces, tracks interpretable behaviors, and captures unique LLM-specific challenges such as spatial misalignment and inconsistent rule-following. Furthermore, leveraging the VGDL, the framework supports the creation of an infinite array of games and levels, either manually or through procedural content generation (PCG)~\cite{khalifa2016general,shaker2016procedural,liu2021deep,guzdial2025procedural}, to rigorously evaluate the capabilities of language models.

\section{Related Work}

Games have recently emerged as powerful benchmarks for evaluating LLM capabilities including reasoning, planning, and spatial understanding \cite{gallotta2024large}. We use \textit{symbolic representation} to refer to ASCII-based structured descriptions of game states where each symbol corresponds to a named entity with defined mechanics, as opposed to \textit{visual representation} (raw pixels or images) or \textit{natural language representation} (free-form text descriptions). 
Several benchmarks rely purely on visual inputs: Atari-GPT~\cite{waytowich2024atari}, VideoGameBench~\cite{zhang2025videogamebench}, MMBench~\cite{liu2024mmbench}, Genie~\cite{bruce2024genie}, NitroGen~\cite{magne2026nitrogen}, and ARC-AGI3~\cite{foundation2026arc}. Others convey game states through natural language wrappers, including BALROG~\cite{paglieri2025balrog}, SmartPlay~\cite{wu2024smartplay}, AgentBench~\cite{liu2024agentbench}, DSGBench~\cite{tang2026dsgbench}, and GTBench~\cite{duan2024gtbench}. ARC-AGI-1, and 2~\cite{chollet2024arc,chollet2025arc} target abstract grid transformation patterns rather than semantically meaningful game entities, making them fundamentally different from inferring entity-level mappings such as \texttt{nokey} $\rightarrow$ \texttt{withkey}.

LMGame-bench~\cite{hu2026lmgamebench} occupies a middle ground: its perception module converts visual inputs into symbolic descriptions, but these are not native engine outputs and game rules are not uniformly encoded. Since symbol mappings are predetermined rather than emerging dynamically, LLMs are not required to infer the meaning of newly appearing symbols. Similarly, GameTraversalBenchmark~\cite{nasir2024gametraversalbenchmark} focuses on path planning in static environments where all entities are given at initialization, requiring no real-time symbol inference or enemy avoidance.
Across all these benchmarks, no work systematically evaluates LLMs' ability to infer the semantics of newly appearing symbols from game rules and act accordingly. GVGAI-LLM addresses this gap: game states are native symbolic representations output directly from the engine, evolving dynamically as gameplay progresses, and LLMs must combine VGDL rules with newly appearing symbols to perform real-time inference, which is a dimension unaddressed by prior work.

\section{GVGAI-LLM: Benchmark Design}

Our benchmark is built upon the GVGAI Java engine, which provides a suite of 118 2D video games designed to evaluate general game-playing agents across a diverse range of tasks and mechanics~\cite{perez2019general,liebana2019general2}. Many of these games are modeled on existing arcade or indie games. While the original framework focused on planning, reactivity, and generalization, its core strength lies in the VGDL~\cite{ebner2013towards,schaul2013video}. Unlike benchmarks restricted to fixed environments~\cite{hu2024games}, VGDL enables the procedural creation of new games and levels to rigorously test agent adaptability, and indeed several game and level generators have been designed that can generate an infinite supply of games~\cite{nielsen2015towards,khalifa2016general,hu2024game}.

To adapt this framework for LLMs, we substantially extended the engine to bridge symbolic game logic with natural language processing (c.f., Section \ref{sec:env}). The environment now exposes both the dynamic game state and the underlying VGDL rules as structured textual representations. These are processed by a prompt generation module, which serializes the environment into a symbolic 2D layout and translates formal mechanics into natural language descriptions.
The resulting prompt, comprising goal descriptions, entity mappings, available actions and coordinate annotations, is passed to the LLM agent. The agent then selects an action based strictly on the current step's information. This action is executed within the environment to complete the closed-loop system. As illustrated in Figure \ref{fig:overall}, this architecture is designed to isolate per-step reasoning and evaluate symbolic generalization independent of external memory or state history.
In addition to standard metrics such as win rate and normalized reward, we define and formulate five interpretable metrics, namely meaningful step ratio, step efficiency and overall score, to provide a comprehensive assessment of agent behavior (c.f., Section \ref{sec:eval-metrics}). The first two metrics are, as far as we can tell, novel.

\subsection{Environment and Agent Design}\label{sec:env}

This section describes the core components responsible for rule and level/scene translation, prompt formatting, agent decision making, and interaction protocols. 



\paragraph{Translator}
\begin{wrapfigure}{r}{0.6\textwidth}
\vspace{-15pt}
  \centering
    \includegraphics[width=1\linewidth]{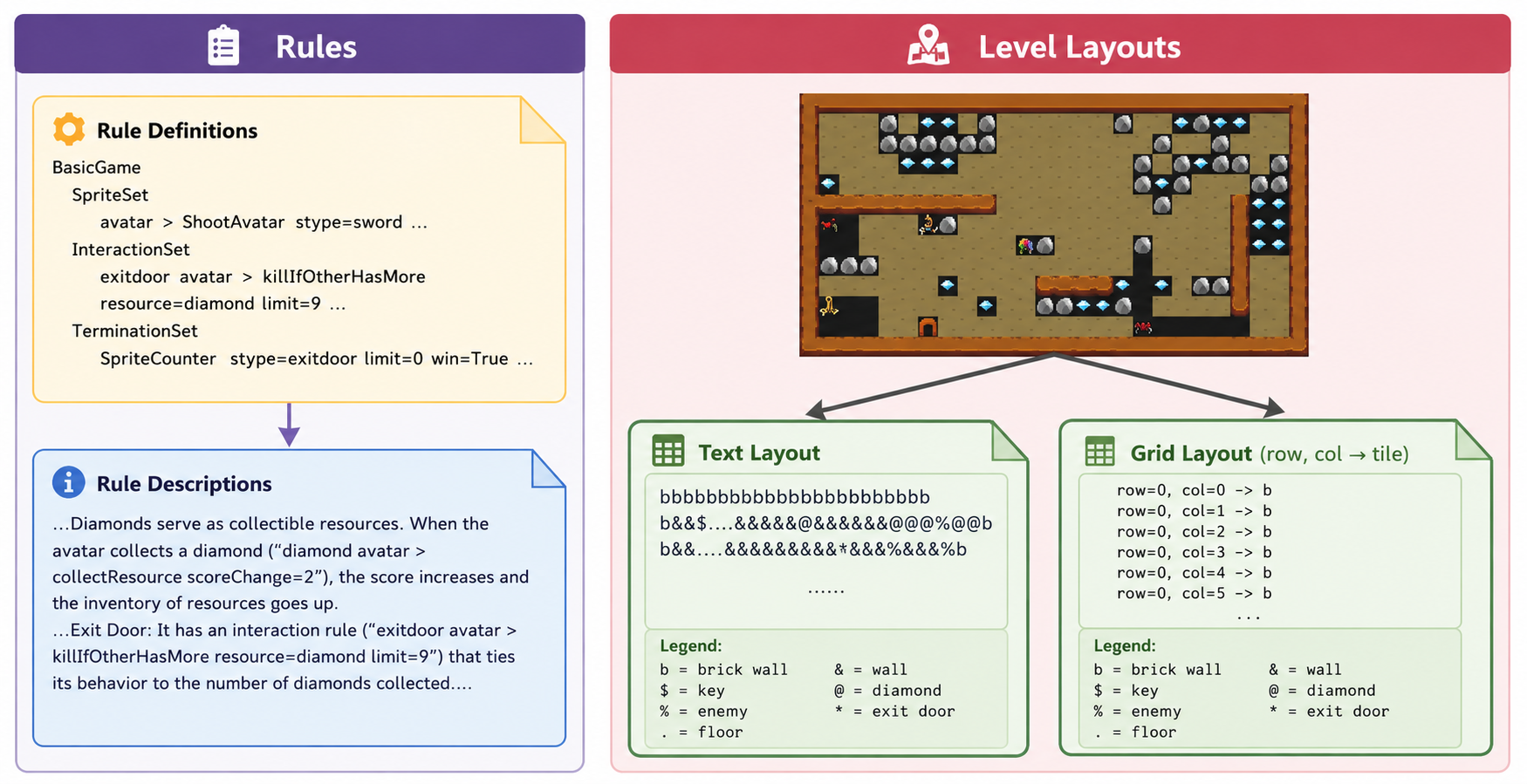}
  \caption{VGDL rule mapping and level layout mapping}
  \label{fig:rule}
  \vspace{-15pt}
\end{wrapfigure}
VGDL~\cite{ebner2013towards,schaul2013video} is a video game description language that encodes game rules and levels/scenes as a compact string of ASCII characters. 
The \textit{Translator} module converts VGDL rule specifications into natural language descriptions. It parses object definitions, interaction rules, and termination conditions from the VGDL script. For example, a rule such as \texttt{avatar key > killSprite} is rendered as: ``\emph{If the avatar touches a key, the key disappears and the avatar obtains it.}'' These textual rule descriptions are provided to downstream modules to support language-based reasoning, as shown in Figure \ref{fig:rule}.

\paragraph{Player}
The \textit{Player} module receives the current game state (represented in ASCII format) along with a high-level strategy. It selects a concrete low-level action (e.g., ``\emph{move right}'') aimed at fulfilling the strategy objective. All decisions are made through language-based reasoning without access to symbolic forward simulation.

\paragraph{Prompt Configuration}
The zero-shot prompt includes several static and dynamic components summarized in Table~\ref{tab:prompt-components}. These elements are presented to the agent at each step to facilitate rule interpretation, spatial reasoning, and action selection. Empirically, we observe that each step consumes approximately 8{,}000 tokens when using reasoning models and around 5{,}000 tokens for non-reasoning models.

\begin{wraptable}{r}{0.5\textwidth}
\vspace{-20pt}
\centering
\caption{Prompt components in \textit{zero-shot} setting.}
\label{tab:prompt-components}
\setlength{\tabcolsep}{10pt}
\begin{tabular}{@{}ll@{}}
\toprule
\textbf{Prompt Component}       & \textbf{Setting} \\
\midrule
Static Rule Description         & Included \\
Action Mapping                  & Included \\
Current Game State              & Included \\
Sprite Mapping                  & Included \\
Avatar Position                 & Included \\
Past States / Past Actions           & Not included \\
\bottomrule
\end{tabular}
\vspace{-12pt}
\end{wraptable}

\paragraph{Zero-shot Interaction Pipeline}

GVGAI-LLM implements a structured, memoryless interaction loop between the environment and the language model. At each discrete time step, the environment serializes the current state into a structured textual prompt, comprising an ASCII spatial map, rule logic, available actions, and the avatar's current position (c.f., Table \ref{tab:prompt-components}). The agent processes this symbolic input and returns a natural language action, which the environment executes to transition the state.


\paragraph{Contextual Prompting}

Although our benchmark adopts a zero-shot setup by default, we also explored a contextual prompting variant in which each input includes a short history, up to 20{,}000 tokens, comprising prior states, actions, and chat. 
However, we found this approach introduced several drawbacks, including compounded reasoning errors, and higher token costs, without significantly improving task performance (as detailed later in Section \ref{sec:contextual}).







\subsection{Evaluation Metrics and Scoring Framework}
\label{sec:eval-metrics}

To assess the behavior of LLM agents in GVGAI-LLM, we introduce a set of evaluation metrics that capture key aspects of decision-making quality. All statistical analyses follow the methodology described in Appendix \ref{app:stats_methods}.

\paragraph{Meaningful Step Ratio}
The \textit{meaningful step ratio} quantifies the proportion of agent actions that produce tangible effects on the environment. This metric captures the agent's ability to engage in purposeful, goal-directed behavior and helps differentiate between effective and redundant actions.
A step is considered \emph{meaningful} if and only if it satisfies
\[
\texttt{isMeaningful(step)} = (\Delta \text{reward} \neq 0) \lor (\Delta \text{state} \neq 0),
\]
where $\Delta \text{state} \neq 0$ indicates that the action causes a visible change in the game world at the symbolic grid level, such as the removal, transformation, or appearance of an entity.
We explicitly exclude moves that trivially cancel each other within a short window---for example, moving left immediately after moving right. We use a window of four steps, which is sufficient to capture immediate back-and-forth oscillations (up to two reversal pairs) while remaining insensitive to deliberate directional changes over longer horizons.
\texttt{ACTION\_NIL} and ineffective actions such as hitting a wall are automatically excluded by the definition.
A high \textit{meaningful step ratio} indicates purposeful, goal-directed behavior; weak agents, regardless of architecture, tend to exhibit low ratios through ineffective patterns such as back-and-forth oscillation or repeated trivial actions. This metric is particularly relevant for evaluating reasoning, planning, and the agent's ability to interact meaningfully with symbolic game environments.

\paragraph{Step Efficiency}
The \textit{step efficiency} evaluates the agent's ability to achieve 
objectives with minimal effort. It is defined as the normalized difference 
between the average number of steps taken to win and the maximum number of 
steps observed across all runs in a level, formulated as:
\[
\texttt{step efficiency} = 1 - \frac{\text{Avg\ Win\ Steps}}{\text{Max\ Taken\ Steps}}.
\]
This value ranges from 0 to 1, with higher values indicating that the agent wins in fewer steps relative to the longest observed episode.

\paragraph{Win Rate} In addition to these, we also track the \textit{win rate}, defined as the fraction of episodes in which the agent successfully completes the level, and total game reward, normalized to the interval \([0, 1]\). While win rate captures binary task completion, normalized reward offers a more granular perspective by measuring partial progress in cases where the agent does not reach the goal.

\paragraph{Normalized Reward} The \textit{normalized reward} is derived via min-max scaling across all experimental trials for every given agent and level:
\[
R = \frac{r - r_{\min}}{r_{\max} - r_{\min} + \epsilon},
\]
where \( r \) is the total reward of a run, \( r_{\min} \) and \( r_{\max} \) are the minimum and maximum rewards observed across all trails, and \( \epsilon \) is a small constant to avoid division by zero.

\paragraph{Overall Score}
To summarize agent behavior across these dimensions, we define an \textit{overall score} by averaging four normalized metrics: meaningful step ratio, total steps (inverted), reward, and win rate.
We deliberately assign equal weights to each metric to ensure balanced evaluation across different aspects of agent behavior, and to maintain interpretability across diverse games. Exploratory tests with manual reweighting yielded no consistent improvements in ranking robustness.
Each component (meaningful step ratio, efficiency, and task success) captures a distinct behavioral property, allowing the final score to reflect multiple dimensions of quality. We avoid assigning manual weights to prevent overfitting to any single behavior type. This formulation supports a robust, interpretable measure for comparing LLM agents across tasks. Since reward distributions vary significantly across GVGAI games, we apply per-game normalization to ensure that the normalized reward captures relative performance within each environment. Pearson correlation analysis (Appendix \ref{app:correlation}) confirms that all four components contribute non-redundant variance (max |r| = 0.62), justifying equal weighting.

\section{Experiments}

We conduct three sets of experiments to evaluate LLM agents within the
\textsc{GVGAI-LLM} framework. (i) We first evaluate \texttt{GPT-4o-mini}
across \emph{all 118 games} under a zero-shot setting to establish a broad
performance baseline and characterize difficulty distribution across the
full game suite, this large-scale sweep is necessary to inform principled
game selection for subsequent experiments. (ii) Building on these findings,
we select six representative games spanning distinct difficulty tiers,
genres, and mechanics, and evaluate nine LLMs under the same zero-shot
setting. (iii) The six LLMs are further compared under a contextual setting.
To ensure a controlled and fair comparison, all models are evaluated using identical prompt templates, input formatting, and decoding parameters.

\def\removed{
\begin{table}[htbp]
\centering
\caption{GPT-4o-mini summary statistics.}
\label{tab:gpt4o_summary}
\setlength{\tabcolsep}{12pt}
\begin{tabular}{lr|lr}
\toprule
\multicolumn{1}{c}{\textbf{Metric}} & \textbf{Value} & \multicolumn{1}{c}{\textbf{Metric}}  & \textbf{Value} \\
\midrule
\#Games (\#Levels) Tested & 118 (540) & Average Meaningful Step Ratio & 49.71\%   \\
\#Zero-Win Levels & 477   & Average Step Efficiency & 0.3293 \\
Overall Win Rate & 10.27\%  & Average Overall Score & 0.2764 \\
\bottomrule
\end{tabular}
\end{table}
}

\subsection{Failure of GPT-4o-mini across 118 Games}\label{sec:full}
\begin{wraptable}{r}{0.48\textwidth}
\vspace{-18pt}
\centering
\caption{GPT-4o-mini summary statistics.}
\label{tab:gpt4o_summary}
\begin{tabular}{lr}
\toprule
\multicolumn{1}{c}{\textbf{Metric}} & \multicolumn{1}{c}{\textbf{Value}} \\
\midrule
\#Games (\#Levels) Tested & 118 (540)  \\
Overall Win Rate & 10.27\% \\
\#Zero-Win Levels & 477   \\
 Average Meaningful Step Ratio & 49.71\%   \\
 Average Step Efficiency & 0.3293 \\
 Average Overall Score & 0.2764 \\
\bottomrule
\end{tabular}
\vspace{-12pt}
\end{wraptable}
We evaluate \texttt{GPT-4o-mini} across all 118 games in the GVGAI-LLM benchmark. Each game is associated with one rule file and up to five distinct level layouts. For each level, we run the model two times using the same underlying game rules, allowing us to assess the model's ability to generalize across structurally different scenarios within the same game. 
Appendix \ref{sec:moregames} provides detailed per-game data. 
According to results summarized in Table~\ref{tab:gpt4o_summary}, \texttt{GPT-4o-mini} fails to complete the vast majority of game levels in GVGAI-LLM, even when the levels are relatively small and conceptually straightforward for humans. These results underscore the challenge posed by GVGAI-LLM and highlight its value in identifying persistent failures in symbolic reasoning and spatial decision-making. 

\subsection{Comparing Nine LLMs under Zero-shot Setting} \label{sec:zero-shot}

Building on the initial findings with \texttt{GPT-4o-mini}, we evaluate nine LLMs under a zero-shot setting (c.f., Table \ref{tab:prompt-components}) on a set of six representative games that span distinct difficulty tiers, genres and mechanics with five levels per game.

\paragraph{Representative Game Selection and Taxonomy}
\texttt{Aliens}, \texttt{Boulderdash}, \texttt{Escape}, \texttt{RealSokoban}, \texttt{Sokoban} and \texttt{Zelda} are used to compare nine LLMs.
To guide this selection, we ranked all 118 games by the average win rate of \texttt{GPT-4o-mini}, spanning the spectrum from zero-win levels to those with perfect success rates. From this ranking, six games were chosen to represent distinct \textit{difficulty tiers} while maintaining diversity in \textit{genre} and \textit{mechanics} as shown in Table \ref{tab:game-taxonomy}. Our selection ranges from real-time action (e.g., \texttt{Aliens}, \texttt{Zelda}) to spatial puzzles (\texttt{Sokoban}, \texttt{RealSokoban}) and open-ended navigation (\texttt{Escape}). \texttt{Boulderdash} is famously hard, because it combines navigation over large levels, a puzzle element (it is possible to get trapped by boulders if you undermine them in the wrong order) and fast-moving enemies. No agent exists that reliably wins all levels, though the best planning agents win more than half. 
Although \texttt{Sokoban} and \texttt{RealSokoban} share the same rules and
similar grid sizes, they differ fundamentally in level origin:
\texttt{Sokoban} levels are procedurally generated within the GVGAI
framework~\cite{perez2019general}, whereas \texttt{RealSokoban} levels are drawn from the XSokoban benchmark, crafted by human designers~\cite{culberson1997sokoban}. This distinction in origin translates directly to puzzle complexity: human-designed levels tend to require significantly longer planning horizons and more careful avoidance of irreversible moves.
This process was informed by established PCG and game AI taxonomies~\cite{shaker2016procedural,Yannakakis2018Artificial}, ensuring a diverse representation~\cite{li2025measuring} of spatiality, interactivity, temporal structure, and difficulty according to the performance of \texttt{GPT-4o-mini} (c.f., Appendix \ref{sec:difficulty-scoring}).

\begin{table}[t]
\centering
\caption{Game taxonomy across temporal granularity, spatiality, genre and level origin.}
\label{tab:game-taxonomy}
\footnotesize
\setlength{\tabcolsep}{4pt}
\renewcommand{\arraystretch}{1.1}
\begin{threeparttable}
\begin{tabular}{llcccccc}
\toprule
\textbf{Dimension} & \textbf{Property} & \textbf{Aliens} & \textbf{Zelda} & \textbf{Escape} & \textbf{Boulderdash} & \textbf{Sokoban} & \textbf{RealSokoban} \\
\midrule
\multirow{2}{*}{Temporal granularity}
  & Real-time    & $\bullet$ & $\bullet$ & $\bullet$ & $\bullet$ & $\circ$ & $\circ$ \\
  & Turn-based   & $\circ$   & $\circ$   & $\circ$   & $\circ$   & $\bullet$ & $\bullet$ \\
\midrule
\multirow{3}{*}{Spatiality}
  & Grid-based           & $\bullet$ & $\bullet$ & $\bullet$ & $\bullet$ & $\bullet$ & $\bullet$ \\
  & Navigation-heavy     & $\circ$   & $\bullet$ & $\bullet$ & $\LEFTcircle$ & $\circ$ & $\circ$ \\
  & Spatial puzzle       & $\circ$   & $\circ$   & $\circ$   & $\LEFTcircle$ & $\bullet$ & $\bullet$ \\
  
\midrule
\multirow{5}{*}{Genre}
  & Shooter / combat         & $\bullet$ & $\circ$ & $\circ$ & $\circ$ & $\circ$ & $\circ$ \\
  & Action-adventure         & $\circ$   & $\bullet$ & $\circ$ & $\circ$ & $\circ$ & $\circ$ \\
  & Open-ended navigation    & $\circ$   & $\circ$ & $\bullet$ & $\circ$ & $\circ$ & $\circ$ \\
  & Real-time puzzle         & $\circ$   & $\circ$ & $\circ$ & $\bullet$ & $\circ$ & $\circ$ \\
  & Spatial puzzle           & $\circ$   & $\circ$ & $\circ$ & $\circ$ & $\bullet$ & $\bullet$ \\
\midrule
\multirow{1}{*}{Level origin}
&Level origin       & P & P & P & P & P & H \\
\bottomrule
\end{tabular}
\begin{tablenotes}
  \footnotesize
  \item $\bullet$ = yes;\quad $\circ$ = no;\quad $\LEFTcircle$ = partial. ~~~~~~~~~~~~~~
  Level origin: P = procedurally generated; H = human-designed.
\end{tablenotes}
\end{threeparttable}
\end{table}

\paragraph{LLM Agents and Setting}
Nine LLMs, \texttt{gpt-4o-mini}, \texttt{o3-mini}, \texttt{gemini-2.0-flash-exp}, \texttt{gemini-2.5-pro}, \texttt{gemini-3-flash}, \texttt{Llama-3.1-405b}, \texttt{Llama-3.2-90b}, \texttt{deepseek-chat-3.2}, and \texttt{deepseek-r3.2}, are evaluated. Each model is run under the same zero-shot prompting setup, with five repeated runs per game level. 

\begin{table*}[htbp]
\centering
\caption{Average win rates of different agents: LLMs in zero-shot, search-based and RL agents.}
\label{tab:algorithm_winrates}
\vspace{2pt}
\resizebox{\textwidth}{!}{%
\begin{tabular}{lcccccc}
\toprule
\textbf{Agent/Model} & \textbf{Aliens} & \textbf{Boulderdash} & \textbf{Escape} & \textbf{RealSokoban} & \textbf{Sokoban} & \textbf{Zelda} \\
\midrule
GPT-4o-mini & 8.0\% & 0.0\% & 4.0\% & 0.0\% & 0.0\% & 0.0\% \\
o3-mini & 80.0\% & 0.0\% & 44.0\% & 0.0\% & 52.0\% & 72.0\% \\
Gemini-2.0-flash & 4.0\% & 0.0\% & 12.0\% & 0.0\% & 0.0\% & 0.0\% \\
Gemini-2.5-pro & 16.0\% & 0.0\% & 36.0\% & 4.0\% & 28.0\% & 8.0\% \\
Gemini-3-flash & 80.0\% & 0.0\% &\textbf{96.0\%} & 4.0\% & \textbf{68.0\%} & \textbf{76.0\%} \\
Llama-3.1-405b & 8.0\% & 0.0\% & 0.0\% & 0.0\% & 0.0\% & 0.0\% \\
Llama-3.2-90b & 0.0\% & 0.0\% & 0.0\% & 0.0\% & 0.0\% & 0.0\% \\
DeepSeek-Chat & 0.0\% & 0.0\% & 24.0\% & 0.0\% & 0.0\% & 36.0\% \\
DeepSeek-r3.2 & 28.0\% & 0.0\% & 60.0\% & \textbf{24.0\%} & 56.0\% & 64.0\% \\
\midrule
olets                & \textbf{100.0\%} & \textbf{56.0\%} & 68.0\% & 0.0\% & 24.0\% & \textbf{76.0\%} \\
sampleMCTS           & \textbf{100.0\%} & 28.0\% & 0.0\%  & 0.0\% & 40.0\% & 24.0\% \\
sampleRHEA           & \textbf{100.0\%} & 4.0\%  & 24.0\% & 0.0\% & 24.0\% & 36.0\% \\
\midrule
baseline3DQN         & 56.0\% & 0.0\%          & 16.0\% & 0.0\% & 0.0\%  & 0.0\%  \\
baseline3PPO         & 64.0\% & 0.0\%          & 12.0\% & 0.0\% & 12.0\% & 8.0\%  \\
\bottomrule
\end{tabular}
}                  
\end{table*}

\paragraph{Results of Multi-model Evaluation} Table \ref{tab:algorithm_winrates} summarizes the performance of nine LLM agents. The results of five search-based and RL-based GVGAI agents, including an international GVGAI competition winner called \texttt{olets}~\cite{perez2019general,liebana2019general2}, are also provided as a reference. Appendices \ref{sec:decision_time} and \ref{sec:rl_config} provide detailed results, including the comparison of decision time.

The results show that the tested LLM agents perform inconsistently across games as shown in Table \ref{tab:algorithm_winrates}. Notably, \texttt{Gemini-3-flash} achieves the strongest overall performance among LLMs across most selected games, achieving 96.0\% in \texttt{Escape} and 68.0\% in \texttt{Sokoban}. Supprisely, \texttt{DeepSeek-r3.2} achieves the highest win rate in \texttt{RealSokoban}, the human designed sokoban levels. In sharp contrast, the \texttt{Llama} models struggle significantly, failing to solve most environments with win rates frequently stagnating at 0.0\%. Notably, all LLM agents score 0.0\% on \texttt{Boulderdash}, suggesting a fundamental challenge beyond current LLM capabilities. Wilson confidence intervals for zero-shot win rates (aggregated over levels) are reported in Appendix \ref{app:wilson}.
Search-based agents (\texttt{olets}, \texttt{sampleMCTS}, \texttt{sampleRHEA}) achieve the strongest overall performance, with all three reaching 100.0\% in \texttt{Aliens} and \texttt{olets} leading in \texttt{Boulderdash} (56.0\%) and \texttt{Zelda} (76.0\%). 
RL-based agents (\texttt{baseline3DQN} and \texttt{baseline3PPO}) perform competitively in action-oriented games, with \texttt{baseline3PPO} achieving 64.0\% in \texttt{Aliens}, already outperforming most LLM agents. 
\begin{figure*}[ht]
    \centering
    \includegraphics[width=1\linewidth]{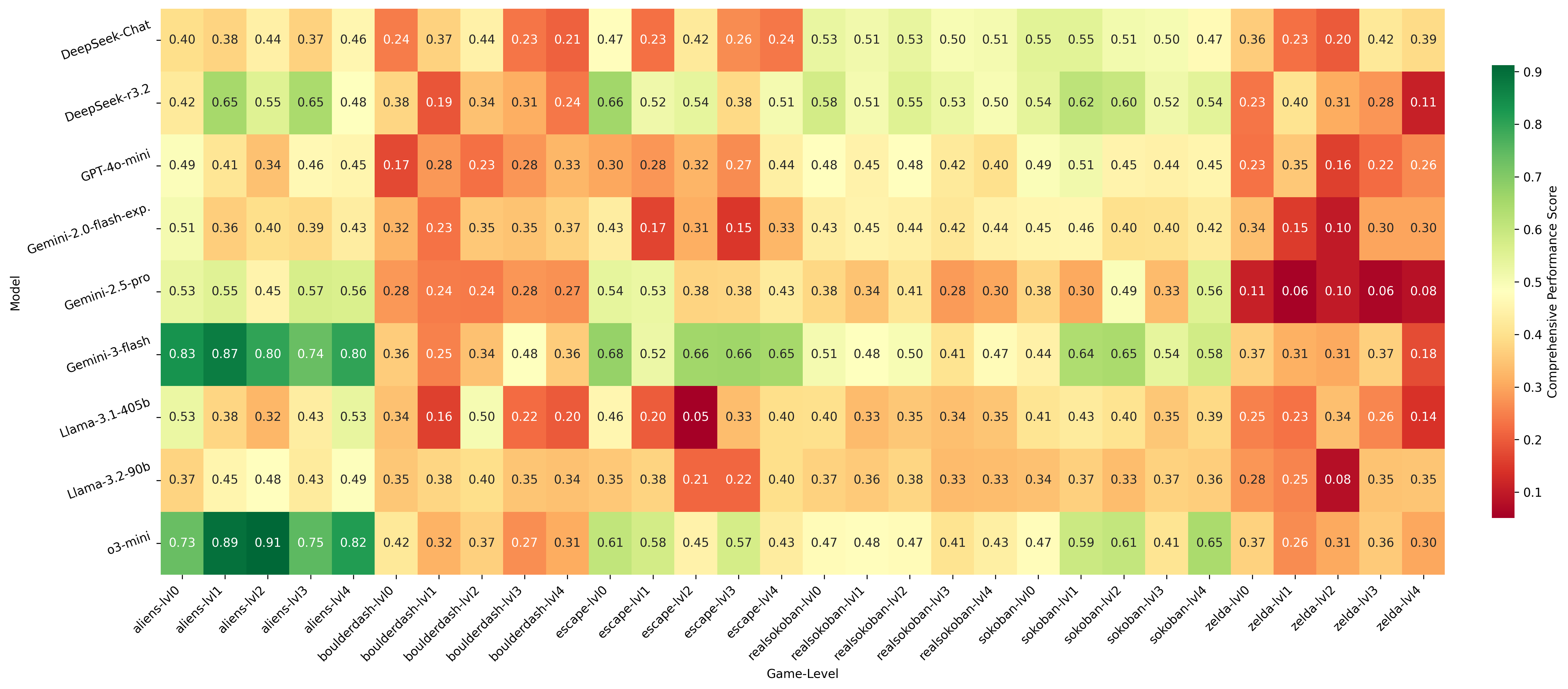}
    \caption{Comprehensive score of each model across six GVGAI games. The comprehensive score aggregates meaningful action ratio, inverse steps, reward and win rate in zero-shot setting.}
    \label{fig:overall-score}
\end{figure*}
\begin{figure*}[ht]
    \centering
    \includegraphics[width=1\linewidth]{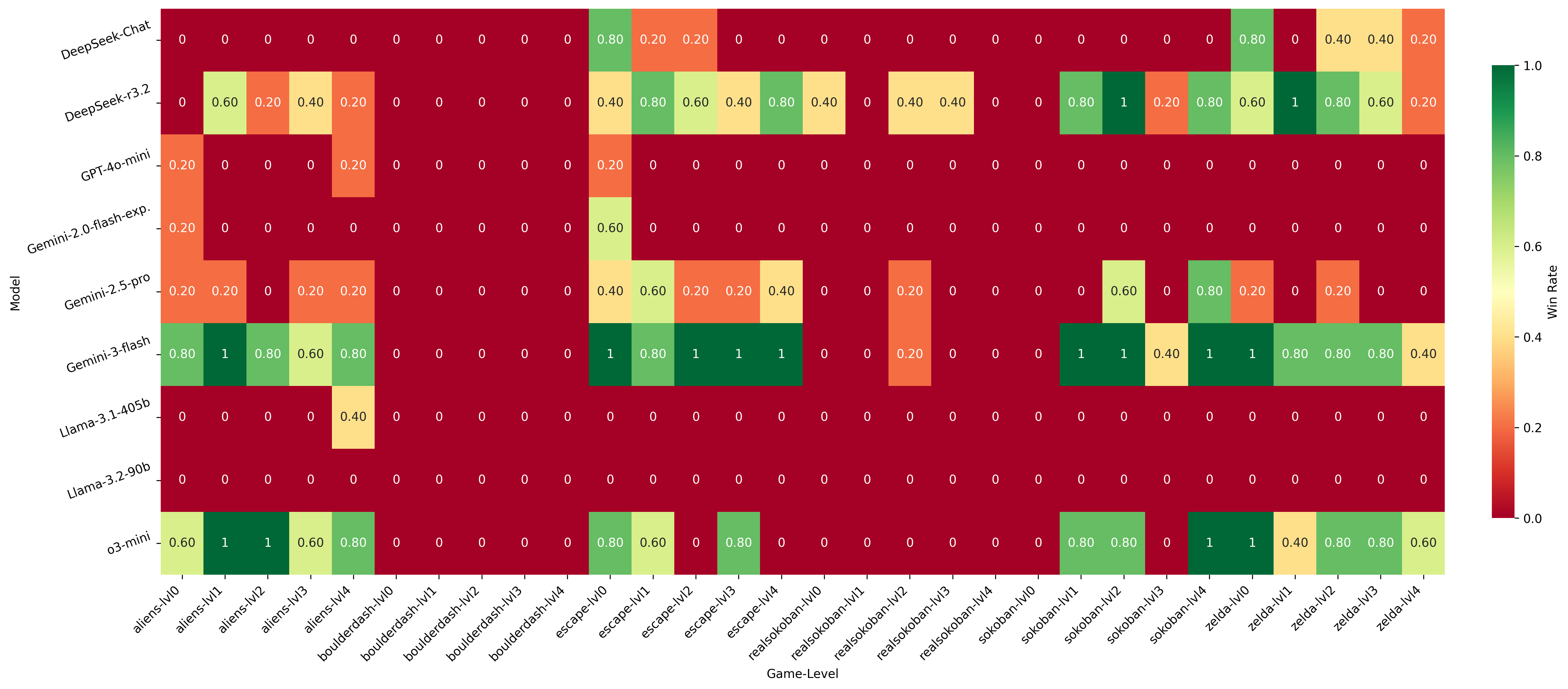}
    \caption{Win rate comparison of nine LLM agents across six GVGAI games in zero-shot setting.}
    \label{fig:win-rate}
\end{figure*}


As shown in Figures \ref{fig:overall-score} and \ref{fig:win-rate},
non-reasoner LLMs generally achieve low comprehensive performance scores
and near-zero win rates across most games. \texttt{GPT-4o-mini} and
\texttt{Gemini-2.0-flash} demonstrate moderate scores in some levels
but rarely succeed in winning. The \texttt{Llama} models perform the
weakest overall, with win rates stagnating at 0.0\% across almost all
games, with the exception of \texttt{Llama-3.1-405b} achieving 40.0\% in
\texttt{aliens} level 4. In contrast, reasoning models exhibit
substantially stronger performance. \texttt{Gemini-3-flash} demonstrates
exceptional consistency, with win rates clustering at 0.8 or 1.0 across
winnable games, except in \texttt{RealSokoban}, where irreversible
decisions prevent recovery from early mistakes. \texttt{o3-mini} achieves
competitive scores across diverse games, particularly excelling in
\texttt{Aliens} and \texttt{Escape}. Overall, reasoner models
substantially outperform standard models in win rate, yet most agents
still fail to win the majority of games across tested levels.
\texttt{Boulderdash} remains largely unsolved across all LLM agents.


\subsection{Comparing Nine LLMs under Contextual Setting} \label{sec:contextual}


\paragraph{Setting}
Six LLMs, \texttt{gpt-4o-mini}, \texttt{o3-mini}, \texttt{gemini-2.0-flash-exp}, \texttt{gemini-2.5-pro}, \texttt{Llama-3.1-405b} and \texttt{Llama-3.2-90b}, are evaluated under a contextual setting, in which agents select actions based on the current symbolic state, integrated chat and state history. Each model is tested with five repeat runs per game level. The \texttt{gemini-3-flash}, \texttt{deepseek-chat} and \texttt{deepseek-r3.2} were excluded from this study due to budgetary constraints. All prompts and decoding parameters are held constant with a temperature of 0.9. Appendix~\ref{sec:overall_result} details results under both prompt settings.


\begin{table*}[htbp]
\centering
\caption{Win rates of key models under the \textit{contextual} prompting setting. 
Bold values indicate results that exceed the corresponding \textit{zero-shot} performance reported in Table~\ref{tab:algorithm_winrates}.}
\label{tab:prompt_variant_comparison}
\vspace{4pt}
\resizebox{\textwidth}{!}{%
\begin{tabular}{lcccccc}
\toprule
\textbf{Model Variant} & \textbf{Aliens} & \textbf{Boulderdash} & \textbf{Escape} & \textbf{RealSokoban} & \textbf{Sokoban} & \textbf{Zelda} \\
\midrule
GPT-4o-mini & 0.0\% & 0.0\% & \textbf{16.0\%} & 0.0\% & 0.0\% & 0.0\% \\

o3-mini  & 52.0\% & 0.0\% & 36.0\% & 0.0\% & 16.0\% & 40.0\% \\

Gemini-2.0-flash  & \textbf{12.0\%} & 0.0\% & 0.0\% & 0.0\% & \textbf{4.0\%} & \textbf{8.0\%} \\

Gemini-2.5-pro  & \textbf{64.0\%} & 0.0\% & 8.0\% & 0.0\% & 24.0\% & \textbf{28.0\%} \\

Llama-3.1-405b & \textbf{32.0\%} & 0.0\% & \textbf{4.0\%} & 0.0\% & 0.0\% & 0.0\% \\

Llama-3.2-90b & \textbf{52.0\%} & 0.0\% & 0.0\% & 0.0\% & 0.0\% & 0.0\% \\
\bottomrule
\end{tabular}%
}
\end{table*}

\paragraph{Zero-shot vs.\ Contextual Settings}
The comparison reveals divergent sensitivities across model families.
Non-thinking models benefit from interaction history:
\texttt{Llama-3.2-90b} achieves 52.0\% in \texttt{Aliens} under the
contextual setting but drops to 0.0\% in zero-shot, suggesting that action-oriented games requiring enemy tracking benefit from temporal context. Similarly, \texttt{Gemini-2.5-pro} and \texttt{Llama-3.1-405b} show strong preferences for contextual prompting in \texttt{Aliens} (64.0\% vs.\ 16.0\% and 32.0\% vs.\ 8.0\%, respectively). Conversely, reasoning-oriented models exhibit an inverse preference: \texttt{o3-mini} improves substantially under zero-shot in planning-heavy games, from 16.0\% to 52.0\% in \texttt{Sokoban} and 40.0\% to 72.0\% in \texttt{Zelda}, consistent with findings that extended context yields negligible or negative gains for reasoning models that already internalize history during chain-of-thought~\cite{leviathan2025prompt}.
Formal significance testing via Fisher's exact test (Appendix \ref{app:fisher}) confirms that no prompt-strategy effect survives BH-FDR correction.


\section{Discussion}

GVGAI-LLM reveals systematic limitations in how current language-only models reason about symbolic, spatially environments. 
We group our findings into three core categories: (1) spatial grounding failures, (2) symbolic identity confusion, and (3) behavioral misalignment.

\paragraph{Spatial Grounding Errors}
Despite receiving structured textual prompts, LLMs frequently misinterpret spatial layouts. First, coordinate confusion, where the model reverses vertical orientation or misaligns row and column positions, often leads to incorrect assumptions about adjacency. Second, hallucinated proximity in sparse layouts causes agents to misjudge distances (e.g., assuming two far-apart objects are adjacent due to horizontal alignment in ASCII maps). These failures often result in ineffective movement and wandering behavior.

We address this issue by introducing two prompt modifications that are applied throughout our experiments. Explicit coordinate tagging removes ambiguity by anchoring object positions numerically (e.g., \texttt{row=3, col=4}). Verbose grounding transforms spatial context into a set of declarative facts (e.g., ``row 2, col 7 is a''). As reported in the results below, these changes improve LLM understanding across games but do not fully resolve the underlying difficulties.
More fundamentally, we observe that LLMs lack the capacity for path planning in the algorithmic sense. Unlike traditional methods such as A$^*$, which search over possible trajectories to compute optimal paths, LLMs operate with limited forward context and no internal simulation of future states. While a reasoning-aware model might theoretically learn to emulate planning through intermediate steps, our outputs do not reveal such structured traces. This limitation aligns with findings from GameTraversalBenchmark~\cite{nasir2024gametraversalbenchmark}, where LLM agents similarly failed to perform multi-step navigation tasks. 

\paragraph{Symbolic Identity Confusion}
A second class of failures concerns the model’s ability to track symbolic identity over time. In many GVGAI games, entity transformations occur through rule-based interactions, for example, an \texttt{avatar} picking up a key transitions from \texttt{nokey} to \texttt{withkey}. 
Despite prompts including both the rule and the updated symbol mapping, some models, such as Gemini-pro, treat \texttt{withkey} as a new and unrelated entity. They often respond as if the avatar has disappeared or cannot act. This failure is not due to prompt ambiguity, as the same information is handled correctly by other models (e.g., GPT-o3-mini), indicating a model-internal reasoning limitation.
Rather than correcting for this behavior via external prompting, GVGAI-LLM treats symbolic identity tracking as an intentional test dimension. 

\paragraph{Behavioral Misalignment}

A third failure class involves misalignment between observed state and chosen action. The most common case is the inappropriate use of \texttt{ACTION\_NIL}, i.e., choosing to do nothing even when meaningful interactions are possible.
In several benchmark games, agents encounter interactive objects such as keys, switches, or portals, yet frequently choose to remain stationary. This behavior reveals a consistent failure to act, even when progression requires explicit movement or interaction. We consider several possible explanations. One possibility is that models assume proximity alone is sufficient to trigger effects, misinterpreting the environment as real-time or timer-driven. Another is that standing still acts as a default behavior when the model is uncertain, reflecting indecision rather than a specific belief about the game. 

Beyond inaction, a related failure involves misdirected action. Such behavioral misalignment persists even when prompts explicitly describe win conditions and available actions. This suggests that the model may struggle to translate symbolic state representations into effective actions. A common failure mode involves incorrect spatial reasoning: for example, 
the agent may continuously walk into a wall that separates it from a key, 
seemingly aware of the goal but unable to plan a valid path around the 
obstacle (quantitatively, losing episodes average 137 consecutive 
wall-bumping steps versus only 14 in winning episodes; Appendix~\ref{app:failure}).
GVGAI-LLM introduced two prompt-level mitigation strategies to mitigate these issues, : (1) \textbf{explicit coordinate tagging}, which anchors entities to precise locations, and (2) \textbf{verbose spatial grounding}, guides the model to explicitly assess what lies in each adjacent direction, helping it reason about possible movement options based on local surroundings.

\paragraph{GVGAI-LLM as a Community Benchmark}
Our results show that even state-of-the-art models fail on the majority
of games, confirming that \textsc{GVGAI-LLM} remains a meaningful and
unsaturated challenge. The persistent failures across spatial reasoning,
symbolic identity tracking, and long-horizon planning suggest that the
benchmark provides substantial headroom for future work. Beyond its
current difficulty, several properties support long-term community
adoption: building on VGDL enables procedural generation of new games
and levels to prevent overfitting as models improve; interpretable
metrics beyond binary win rate allow fine-grained behavioral comparison;
and support for symbolic, visual, and multimodal interfaces makes the
benchmark accessible across a broad range of model families without
specialized infrastructure.

\section{Conclusion}
We introduce GVGAI-LLM, a benchmark that reuses and extends the classic GVGAI environment for LLM agents. The benchmark supports symbolic, visual, and multimodal interfaces, defines interpretable metrics, and ensures reproducibility. It offers a compact setting for analyzing how language models make decisions in games. Our experiments show that most models fail on most games, even when the environments are small. 
This reveals fundamental weaknesses among frontier models in symbolic reasoning, spatial understanding, and planning. GVGAI-LLM remains a difficult challenge and can support further study of LLM behavior in structured decision problems. 
In the future, we plan to extend GVGAI-LLM to support language models that not only play games, but also design them. This includes using LLMs to generate new game rules and levels, which can then serve as novel test cases for evaluating other models, shifting the focus from agentic behavior to creative and generative capabilities. These design tasks may also provide a new lens on how LLMs reason about rules and game mechanisms, rather than just executing them.

\bibliographystyle{unsrtnat}

\bibliography{ijcai25}
\newpage
\clearpage
\appendix


\section{ChatGPT-4o-mini on 118 games}\label{sec:moregames}

Our code is available at: \url{https://anonymous.4open.science/r/GVGAI_jpype-00DA/README.md}

To further examine the generalization ability of \texttt{GPT-4o-mini}, we report detailed results across a broader set of games in the GVGAI-LLM benchmark. Tables~\ref{tab:gpt4o_results_part1} and~\ref{tab:gpt4o_results_part2} present the model's performance on 118 environments, covering meaningful step ratio, win rate, overall behavioral score, and step efficiency. While the model demonstrates high engagement in many environments (e.g., \texttt{Flower}, \texttt{Infection}, \texttt{Butterflies}), it achieves consistently low win rates, particularly in puzzle and planning-heavy games.

We also assess the effect of coordinate-based spatial grounding. Table~\ref{tab:coord-tagger-winrate} compares win rates with and without coordinate taggers across six spatially demanding games. Although slight improvements are observed in \texttt{aliens} and \texttt{escape}, none reach statistical significance under Fisher's exact test, suggesting that coordinate tagging alone is insufficient to resolve core spatial reasoning limitations.

\begin{table}[htbp]
\centering
\caption{Winrate without coordinate tagger vs. with coordinate tagger of ChatGPT-4o-mini. The significance of any improvement is judged using Fisher's exact test over 25 episodes per game. None of the improvements were statistically significant.}\label{tab:coord-tagger-winrate}
\begin{tabular}{lcc}
\toprule
\textbf{Game} & \textbf{Without tagger} & \textbf{With tagger} \\
\midrule
aliens       & 0.00 & 0.08 \\
boulderdash  & 0.00 & 0.00 \\
escape       & 0.00 & 0.04 \\
realsokoban  & 0.00 & 0.00 \\
sokoban      & 0.00 & 0.00 \\
zelda        & 0.00 & 0.00 \\
\bottomrule
\end{tabular}
\end{table}
\newpage
\clearpage

\begin{table}[htbp]
\centering

\caption{GPT-4o Performance - Part 1}\label{tab:gpt4o_results_part1}
\begin{tabular}{p{2.5cm}@{\hspace{0.3cm}}*{4}{c}}

\toprule
\textbf{Game} & \textbf{M.Ratio} & \textbf{Win\%} & \textbf{Score} & \textbf{Eff.} \\
\midrule
Aliens & 51.8 & 0.0 & 0.213 & 0.705 \\
Angelsdemons & 69.8 & 0.0 & 0.285 & 0.857 \\
Assemblyline & 56.1 & 0.0 & 0.364 & 0.137 \\
Avoidgeorge & 77.9 & 0.0 & 0.257 & 0.755 \\
Bait & 9.6 & 12.5 & 0.133 & 0.000 \\
Beltmanager & 48.5 & 0.0 & 0.174 & 0.611 \\
Blacksmoke & 82.1 & 0.0 & 0.277 & 0.136 \\
Boloadventures & 48.7 & 0.0 & 0.326 & 0.633 \\
Bomber & 40.2 & 0.0 & 0.155 & 0.114 \\
Bomberman & 86.4 & 0.0 & 0.255 & 0.000 \\
Boulderchase & 51.7 & 0.0 & 0.185 & 0.175 \\
Boulderdash & 19.9 & 0.0 & 0.213 & 0.159 \\
Brainman & 62.9 & 0.0 & 0.450 & 0.476 \\
Bravekeeper & 72.3 & 0.0 & 0.242 & 0.827 \\
Butterflies & 76.4 & 80.0 & 0.512 & 0.791 \\
Cakybaky & 72.4 & 0.0 & 0.225 & 0.781 \\
Camelrace & 53.9 & 20.0 & 0.232 & 0.000 \\
Catapults & 0.0 & 0.0 & 0.040 & 0.000 \\
Cec1 & 31.6 & 0.0 & 0.117 & 0.174 \\
Cec2 & 23.3 & 0.0 & 0.219 & 0.188 \\
Cec3 & 58.7 & 100.0 & 0.443 & 0.860 \\
Chainreaction & 67.0 & 0.0 & 0.244 & 0.844 \\
Chase & 55.9 & 0.0 & 0.391 & 0.776 \\
Chipschallenge & 58.7 & 0.0 & 0.236 & 0.801 \\
Chopper & 79.1 & 0.0 & 0.264 & 0.787 \\
Clusters & 72.8 & 0.0 & 0.227 & 0.860 \\
Colourescape & 3.7 & 0.0 & 0.157 & 0.000 \\
Cookmepasta & 31.7 & 0.0 & 0.366 & 0.478 \\
Cops & 47.0 & 0.0 & 0.391 & 0.665 \\
Crossfire & 82.2 & 0.0 & 0.297 & 0.644 \\
Decepticoins & 50.1 & 60.0 & 0.320 & 0.809 \\
Deceptizelda & 32.6 & 100.0 & 0.374 & 0.000 \\
Defem & 62.5 & 0.0 & 0.199 & 0.980 \\
Defender & 79.4 & 60.0 & 0.461 & 0.728 \\
Deflection & 0.0 & 0.0 & 0.037 & 0.510 \\
Digdug & 32.2 & 0.0 & 0.194 & 0.175 \\
Donkeykong & 59.6 & 0.0 & 0.266 & 0.045 \\
Doorkoban & 40.7 & 0.0 & 0.263 & 0.630 \\
Dungeon & 78.6 & 0.0 & 0.390 & 0.892 \\
Eggomania & 57.5 & 0.0 & 0.190 & 0.739 \\
Eighthpassenger & 66.9 & 0.0 & 0.219 & 0.000 \\
Enemycitadel & 68.9 & 0.0 & 0.416 & 0.525 \\
Escape & 44.9 & 0.0 & 0.179 & 0.566 \\
Factorymanager & 47.9 & 100.0 & 0.504 & 0.238 \\
Firecaster & 14.2 & 0.0 & 0.324 & 0.136 \\
Fireman & 55.8 & 0.0 & 0.254 & 0.160 \\
Firestorms & 72.4 & 0.0 & 0.294 & 0.737 \\
Flower & 71.6 & 100.0 & 0.868 & 0.980 \\
Freeway & 90.6 & 0.0 & 0.275 & 0.000 \\
Frogs & 52.7 & 0.0 & 0.173 & 0.493 \\
Greedymouse & 14.5 & 0.0 & 0.080 & 0.081 \\
Gymkhana & 19.4 & 0.0 & 0.089 & 0.000 \\
Hungrybirds & 27.9 & 0.0 & 0.209 & 0.463 \\
Iceandfire & 38.4 & 0.0 & 0.156 & 0.602 \\
Ikaruga & 62.4 & 0.0 & 0.206 & 0.000 \\
Infection & 96.3 & 80.0 & 0.607 & 0.000 \\
Intersection & 78.1 & 0.0 & 0.251 & 0.867 \\
Invest & 60.5 & 0.0 & 0.201 & 0.975 \\
\bottomrule

\end{tabular}

\end{table}

\begin{table}[htbp]
\centering
\caption{GPT-4o Performance - Part 2}\label{tab:gpt4o_results_part2}

\begin{tabular}{p{2.2cm}@{\hspace{0.2cm}}*{4}{c}}
\toprule
\textbf{Game} & \textbf{M.Ratio} & \textbf{Win\%} & \textbf{Score} & \textbf{Eff.} \\
\midrule

Investdie & 69.6 & 0.0 & 0.219 & 0.982 \\
Islands & 36.1 & 0.0 & 0.127 & 0.361 \\
Jaws & 50.1 & 0.0 & 0.174 & 0.244 \\
Killbillvol1 & 12.6 & 0.0 & 0.277 & 0.000 \\
Labyrinth & 39.3 & 0.0 & 0.206 & 0.632 \\
Labyrinthdual & 0.0 & 0.0 & 0.135 & 0.000 \\
Lasers & 56.3 & 0.0 & 0.183 & 0.221 \\
Lasers2 & 43.8 & 0.0 & 0.222 & 0.183 \\
Lemmings & 28.6 & 0.0 & 0.334 & 0.000 \\
Mirrors & 22.2 & 0.0 & 0.184 & 0.000 \\
Missilecommand & 62.6 & 40.0 & 0.306 & 0.000 \\
Modality & 15.9 & 20.0 & 0.330 & 0.000 \\
Overload & 28.4 & 0.0 & 0.323 & 0.000 \\
Pacman & 0.0 & 0.0 & 0.048 & 0.000 \\
Painter & 42.1 & 60.0 & 0.429 & 0.000 \\
Plants & 93.1 & 0.0 & 0.318 & 0.000 \\
Plaqueattack & 84.1 & 20.0 & 0.345 & 0.000 \\
Pokemon & 48.5 & 80.0 & 0.462 & 0.000 \\
Portals & 83.7 & 0.0 & 0.253 & 0.000 \\
Racebet & 39.6 & 0.0 & 0.153 & 0.000 \\
Racebet2 & 51.5 & 20.0 & 0.235 & 0.000 \\
Realportals & 29.6 & 0.0 & 0.115 & 0.000 \\
Realsokoban & 34.7 & 0.0 & 0.372 & 0.000 \\
Rivers & 23.5 & 0.0 & 0.250 & 0.000 \\
Roadfighter & 83.9 & 0.0 & 0.251 & 0.000 \\
Roguelike & 38.8 & 0.0 & 0.199 & 0.000 \\
Run & 72.6 & 0.0 & 0.231 & 0.000 \\
Seaquest & 84.8 & 0.0 & 0.283 & 0.000 \\
Sheriff & 76.4 & 0.0 & 0.247 & 0.000 \\
Shipwreck & 37.4 & 40.0 & 0.296 & 0.000 \\
Sistersavior & 51.0 & 50.0 & 0.297 & 0.000 \\
Sokoban & 44.7 & 0.0 & 0.398 & 0.000 \\
Solarfox & 50.9 & 0.0 & 0.177 & 0.000 \\
Superman & 71.2 & 0.0 & 0.241 & 0.000 \\
Surround & 22.1 & 100.0 & 0.349 & 0.000 \\
Survivezombies & 77.1 & 20.0 & 0.324 & 0.000 \\
Tercio & 9.3 & 0.0 & 0.183 & 0.000 \\
Thecitadel & 39.0 & 0.0 & 0.383 & 0.000 \\
Themole & 56.4 & 0.0 & 0.257 & 0.000 \\
Theshepherd & 25.0 & 0.0 & 0.348 & 0.000 \\
Thesnowman & 36.2 & 0.0 & 0.327 & 0.000 \\
Trappedhero & 0.0 & 0.0 & 0.185 & 0.000 \\
Treasurekeeper & 39.4 & 0.0 & 0.148 & 0.000 \\
Vortex & 51.2 & 0.0 & 0.289 & 0.000 \\
Waferthinmints & 64.2 & 0.0 & 0.205 & 0.000 \\
Waferthinmin... & 35.3 & 100.0 & 0.376 & 0.000 \\
Waitforbreak... & 13.3 & 0.0 & 0.075 & 0.000 \\
Watergame & 39.1 & 0.0 & 0.138 & 0.000 \\
Waterpuzzle & 0.0 & 0.0 & 0.222 & 0.000 \\
Waves & 70.2 & 0.0 & 0.240 & 0.000 \\
Whackamole & 54.3 & 20.0 & 0.252 & 0.000 \\
Wildgunman & 71.3 & 0.0 & 0.233 & 0.000 \\
Witnessprote... & 98.8 & 0.0 & 0.352 & 0.000 \\
Witnessprote... & 73.2 & 0.0 & 0.241 & 0.000 \\
Wrapsokoban & 29.2 & 0.0 & 0.359 & 0.000 \\
X-Racer & 94.5 & 0.0 & 0.297 & 0.078 \\
Zelda & 9.7 & 0.0 & 0.083 & 0.000 \\
Zenpuzzle & 31.6 & 0.0 & 0.364 & 0.000 \\
\bottomrule
\end{tabular}
\end{table}

\newpage
\clearpage

\section{Zero-Shot Prompt Example}

We design a structured prompt to support zero-shot reasoning in symbolic games. The prompt contains multiple components, including natural language descriptions of the game rules, mechanics analysis, available actions, strategy advice, and a fully verbalised game state. An excerpt of the prompt is shown below (abridged for space):

\begin{quote}
\texttt{
User\\
=== Game Rules ===\\
Game rules in natural language:\\
\\
\# Game Analysis\\
\\
\textbf{Genre:} This is a puzzle-adventure game with elements of strategy and item collection.\\
\\
\textbf{Mechanics:}\\
1. \textbf{Sprites:} The game includes avatars, walls, doors, gems, keys, and exits.\\
2. \textbf{Transformation and Interaction:} The avatar collects keys, avoids enemies, and triggers effects via sprite interactions.\\
3. \textbf{Win/Loss:} Reaching the exit wins the game. Losing occurs if the avatar is destroyed.\\
\\
\textbf{Strategy Suggestions:}\\
- Collect keys to unlock doors and reach the exit.\\
- Avoid enemies (gems) unless necessary.\\
- Use missile keys strategically.\\
\\
=== Available Actions ===\\
0: ACTION\_NIL\\
1: ACTION\_LEFT\\
2: ACTION\_RIGHT\\
3: ACTION\_DOWN\\
4: ACTION\_UP\\
\\
=== Important Mechanics Notice ===\\
Some directional actions may rotate the avatar without movement. Repeating the direction may be needed. Avoid null actions. Interpret the state carefully and act meaningfully.\\
\\
=== Sprite Mapping ===\\
avatar $\rightarrow$ \texttt{'a'}\\
background floor $\rightarrow$ \texttt{'.'}\\
exit $\rightarrow$ \texttt{'e'}\\
diamond $\rightarrow$ \texttt{'\%'}\\
wall $\rightarrow$ \texttt{'b'}\\
dirt $\rightarrow$ \texttt{'\&'}\\
enemy $\rightarrow$ \texttt{'\$'}\\
\\
=== Current State ===\\
bbbbbbbbbbbbbbbbbbbbbbbbbb\\b\&@@\&\&\&\&\&@@@........a....b\\b\&\%\&\&b@b\&\&\&\%\&\&.\&\&\&@@\&\&\&bbb\\
...\\
Each line shows entity at (row, col).\\
row=1, col=3 $\rightarrow$ \texttt{a}\\
row=2, col=7 $\rightarrow$ \texttt{\%} (diamond)\\
row=2, col=8 $\rightarrow$ \texttt{\&} (dirt)\\
row=3, col=2 $\rightarrow$ \texttt{\%} (diamond)\\
row=1, col=14 $\rightarrow$ \texttt{\$} (enemy)\\
...
}
\end{quote}

The agent is instructed to output a response in the format:
\texttt{\textbackslash\textbackslash\ Action:<action number>}\\
It also provides a one-line justification of the action in relation to the current strategy.

This structured prompt enables the LLM to (1) understand symbolic state representations, (2) reason over spatial layout and interaction rules, and (3) produce meaningful, goal-aligned actions in a zero-shot setting.

\section{Game Difficulty Scoring and Labeling}
\label{sec:difficulty-scoring}

In addition to measuring agent quality, we also assign difficulty labels to games based on how well agents perform. This is useful for identifying challenging scenarios and characterizing environment complexity.

We define a difficulty score for each game by aggregating performance indicators across all evaluated agents. The score combines win rate, normalized reward, and step efficiency using a weighted sum:

\begin{align}
\text{Difficulty Score} =\ & 0.6 \cdot \text{Win Rate} \notag \\
                           & + 0.2 \cdot \text{Normalized Reward} \notag \\
                           & + 0.2 \cdot \text{Step Efficiency}
\end{align}

This formulation gives higher priority to task success, while still rewarding agents that make meaningful progress or act efficiently. Based on the resulting score, we assign each game a difficulty label according to the following criteria:

\begin{table}[h]
\centering
\caption{Difficulty label assignment based on aggregated agent scores}
\begin{tabular}{ll}
\toprule
\text{Score Range} & \text{Difficulty Label} \\\midrule
$\geq 0.8$        & very\_easy \\
$0.6 - 0.8$       & easy       \\
$0.4 - 0.6$       & medium     \\
$0.2 - 0.4$       & hard       \\
$< 0.2$           & very\_hard \\
No agent succeeded & unbeaten \\
\bottomrule
\end{tabular}

\end{table}

This classification system provides a standardized framework for analyzing task complexity. It allows researchers to group games by difficulty, diagnose failure modes in harder environments, and prioritize settings for testing more advanced prompting or agent architectures.

Together, the overall agent score and the difficulty labeling mechanism offer a complete behavioral evaluation pipeline, enabling systematic study of both agent performance and environment characteristics within GVGAI-LLM.

\section{Decision Time Comparison Between LLM Agents and MCTS} \label{sec:decision_time}

Table~\ref{tab:decision} reports the per-move latency for all LLM agents evaluated in GVGAI-LLM. For models served via the Portkey API, we use the median (p50) latency statistics collected over the past month. For DeepSeek models, latency is measured as the average response time over 10 representative game prompts. In contrast, MCTS is not an API-based model but a symbolic planning algorithm that directly simulates game rules; its decision time is set to the standard per-move time budget (0.04s) defined in the original GVGAI competition.

The results reveal that LLM agents are two to three orders of magnitude slower than symbolic search methods. While real-time playability is not the focus of our benchmark, this latency gap highlights the computational inefficiency of current LLM-based agents compared to classical planning algorithms.

\begin{table}[htbp]
\centering
\caption{Estimated decision time per move (in seconds) for LLM agents and MCTS.} \label{tab:decision}
\begin{tabular}{lcc}
\toprule
\textbf{Agent} & \textbf{Model} & \textbf{Time/Move} \\
\midrule
Gemini-2.5-pro\textsuperscript{*}        & LLM     & 27.82 \\
Gemini-2.0-flash-exp\textsuperscript{*}  & LLM     & 1.05  \\
GPT-o3-mini\textsuperscript{*}           & LLM     & 35.94 \\
GPT-4o-mini\textsuperscript{*}           & LLM     & 2.37  \\
DeepSeek Reasoner\textsuperscript{†}     & LLM     & 166.00 \\
DeepSeek Chat\textsuperscript{†}         & LLM     & 1.80  \\
MCTS\textsuperscript{‡}                  & Search Algorithm  & 0.04  \\
\bottomrule

\end{tabular}

\vspace{0.5em}
\raggedright
\textsuperscript{*} Served via Portkey API; Median (p50) latency from Portkey API over the past month. \\
\textsuperscript{†} Latency averaged over prompts from 10 GVGAI games. \\
\textsuperscript{‡} Monte Carlo Tree Search (MCTS) agent using symbolic rule simulation; 0.04s is the per-move time budget defined in the original GVGAI competition settings.
\end{table}

\section{RL Agent Configuration Details}
\label{sec:rl_config}
To compare LLM agents with reinforcement learning (RL) baselines, we trained PPO and DQN agents using standard configurations from the Stable Baselines3 library, with minimal adjustments for symbolic grid-based input. Specifically, both policies employ a custom feature extractor (\texttt{PaddedEmbeddingCNN}) designed to handle discrete int8 observations. All hyperparameters are fixed and shared across games and levels. No tuning was performed, in order to ensure a fair comparison with LLM agents, which are also evaluated in a zero-shot manner.

\vspace{1em}
\noindent \textbf{PPO Configuration (used without tuning)}
\begin{itemize}
  \item \textbf{Policy}: \texttt{MlpPolicy} with custom extractor (\texttt{PaddedEmbeddingCNN})
  \item \textbf{Embedding dimension}: 32
  \item \textbf{Feature dimension}: 256
  \item \textbf{MLP architecture}:
    \begin{itemize}
      \item $\pi$-network: [128, 64]
      \item Value-network: [128, 64]
    \end{itemize}
  \item \textbf{Learning rate}: 3e-4
  \item \textbf{n\_steps}: 2048
  \item \textbf{Batch size}: 256
  \item \textbf{n\_epochs}: 10
  \item \textbf{Discount factor} ($\gamma$): 0.99
  \item \textbf{GAE lambda}: 0.95
  \item \textbf{Entropy coefficient}: 0.01
  \item \textbf{Rollout buffer}: \texttt{Int8RolloutBuffer}
  \item \textbf{Device}: RTX4090
\end{itemize}

\vspace{1em}
\noindent \textbf{DQN Configuration (used without tuning)}
\begin{itemize}
  \item \textbf{Policy}: \texttt{MlpPolicy} with custom extractor (\texttt{PaddedEmbeddingCNN})
  \item \textbf{Embedding dimension}: 32
  \item \textbf{Feature dimension}: 256
  \item \textbf{Q-network architecture}: [128, 64]
  \item \textbf{Learning rate}: 3e-4
  \item \textbf{Replay buffer}: \texttt{Int8ReplayBuffer}
  \item \textbf{Buffer size}: $\min(1{,}000{,}000,\ \text{total timesteps})$
  \item \textbf{Batch size}: 256
  \item \textbf{Discount factor} ($\gamma$): 0.99
  \item \textbf{Exploration}:
    \begin{itemize}
      \item Initial epsilon: 1.0
      \item Final epsilon: 0.02
      \item Exploration fraction: 0.1
    \end{itemize}
  \item \textbf{Target update interval}: 1000
  \item \textbf{Device}: RTX4090
\end{itemize}
\vspace{1em}
\noindent \textbf{Evaluation Protocol:}

All agents, including reinforcement learning (PPO, DQN), search-based (MCTS), and language model agents, evaluated on six GVGAI games: \texttt{zelda}, \texttt{aliens}, \texttt{boulderdash}, \texttt{escape}, \texttt{sokoban}, and \texttt{realsokoban} are under the same experimental protocol. Each game contains 5 levels, resulting in 30 unique game-level combinations. For every level, each agent is evaluated across 5 independent episodes. All reported metrics (e.g., win rate, total reward, number of steps) are averaged over these 5 runs per level, ensuring consistent comparison across models and methods. Heatmaps for LLM win rate are presented separately on each level.

\section{Additional Results by Prompt Setting}
\label{sec:overall_result}
\subsection{Zero-shot Prompt}

\begin{figure*}[htbp]
  \centering
  \includegraphics[width=\textwidth]{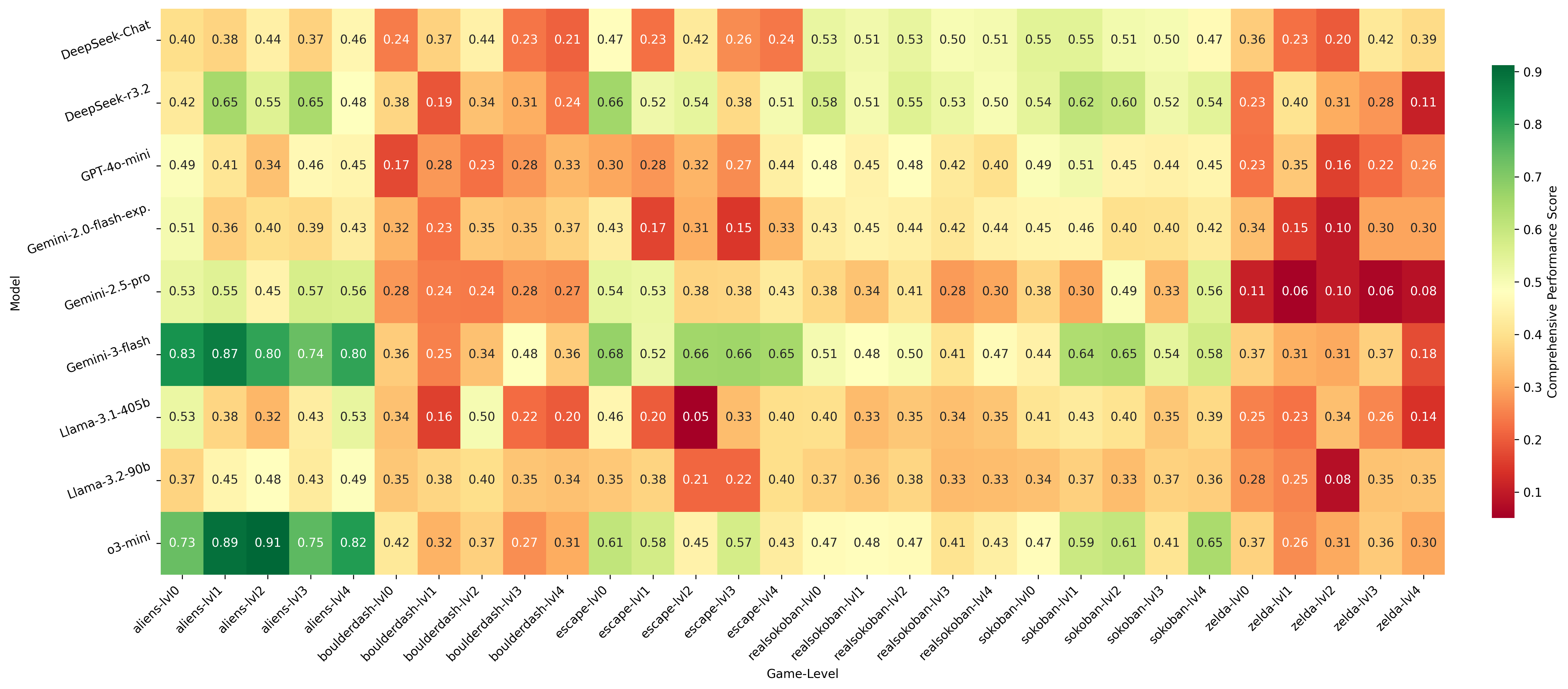}
  \caption{Zero-shot prompt: comprehensive performance heatmap.}
  \label{fig:app:zeroshot:comprehensive}
\end{figure*}

The following figures present the full evaluation results under the zero-shot prompt setting, where agents receive no in-context examples or prior game history. The comprehensive performance heatmap (Figure~\ref{fig:app:zeroshot:comprehensive}) summarizes overall behavioral scores across all agent-game combinations. Subsequent heatmaps report win rate, normalized reward, meaningful step ratio, and total steps respectively, offering a fine-grained view of agent behavior in the absence of contextual guidance.

\begin{figure*}[htbp]
  \centering
  \includegraphics[width=\textwidth]{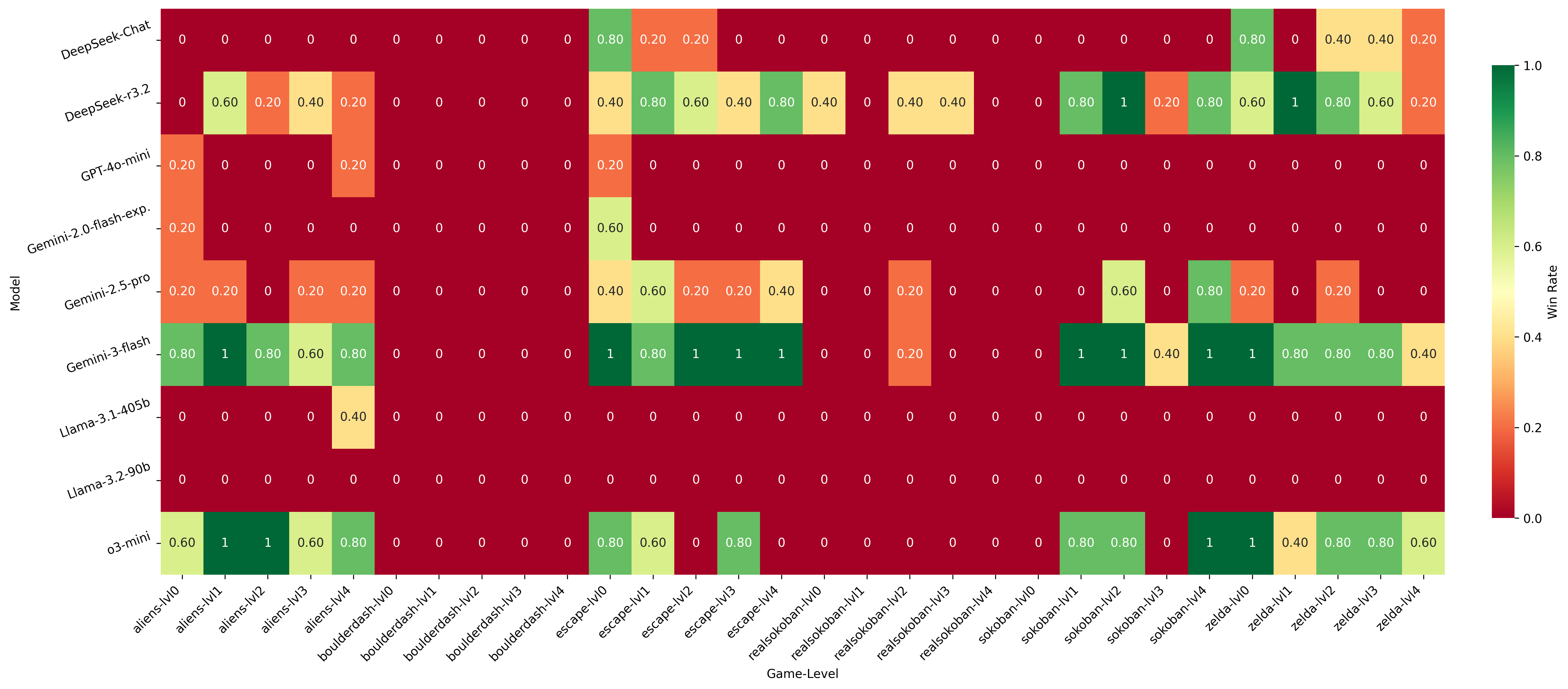}
  \caption{Zero-shot prompt: win rate heatmap.}
\end{figure*}

\begin{figure*}[htbp]
  \centering
  \includegraphics[width=\textwidth]{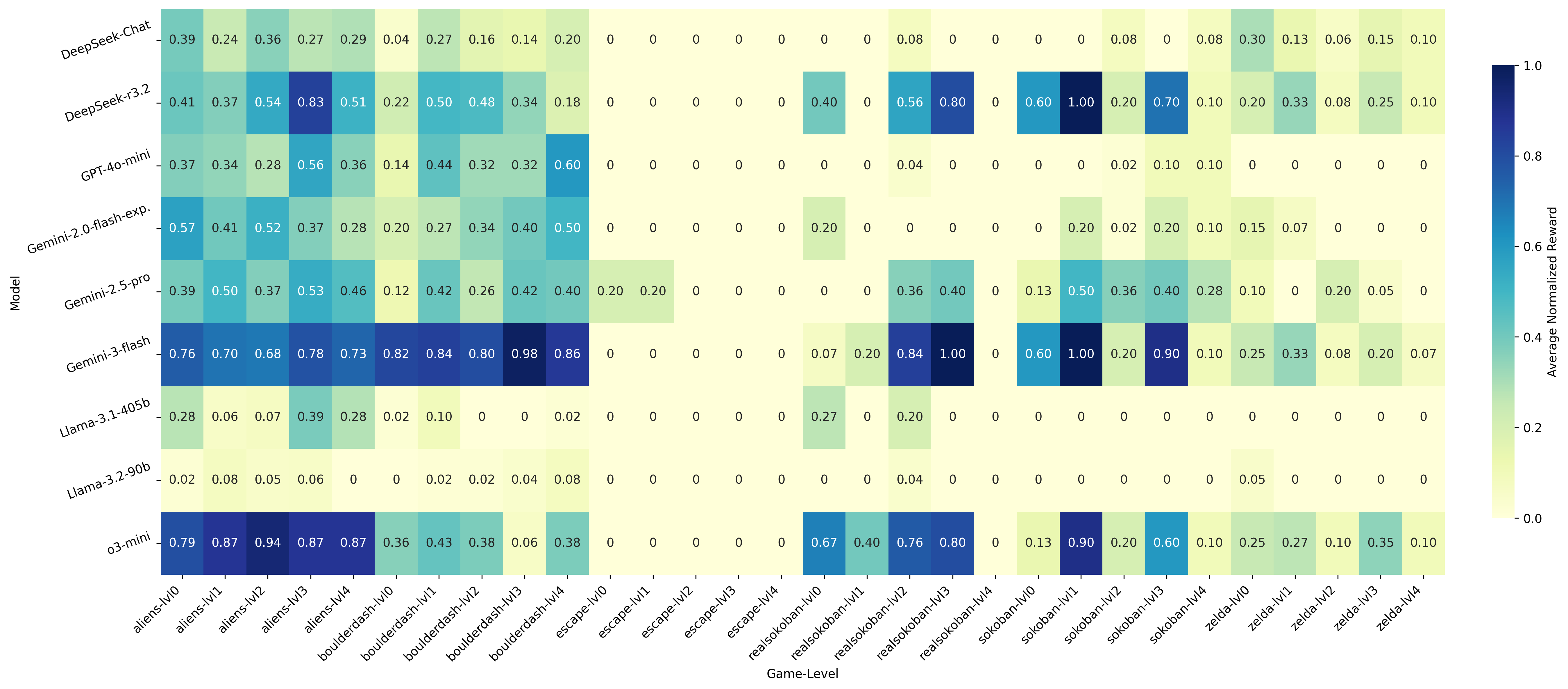}
  \caption{Zero-shot prompt: normalized reward heatmap.}
\end{figure*}

\begin{figure*}[htbp]
  \centering
  \includegraphics[width=\textwidth]{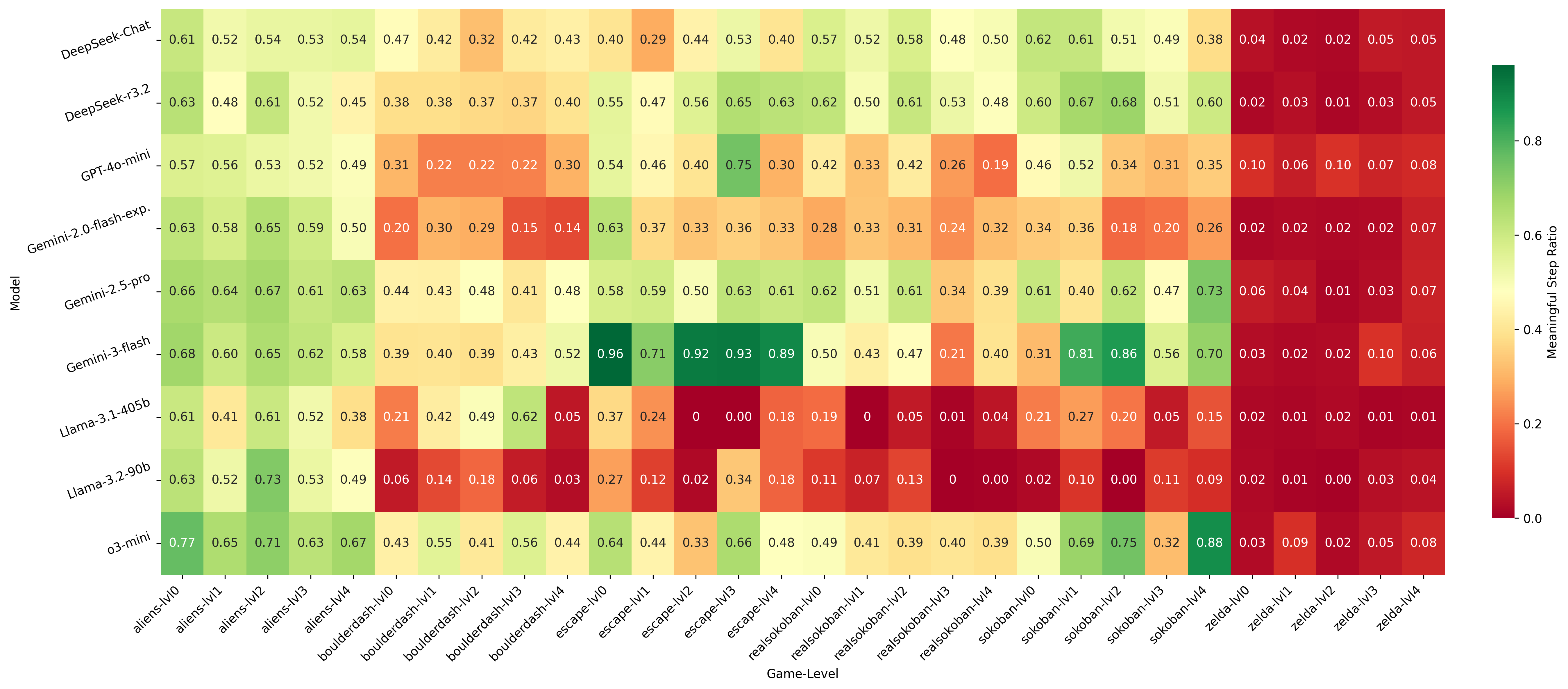}
  \caption{Zero-shot prompt: meaningful step ratio heatmap.}
\end{figure*}

\begin{figure*}[htbp]
  \centering
  \includegraphics[width=\textwidth]{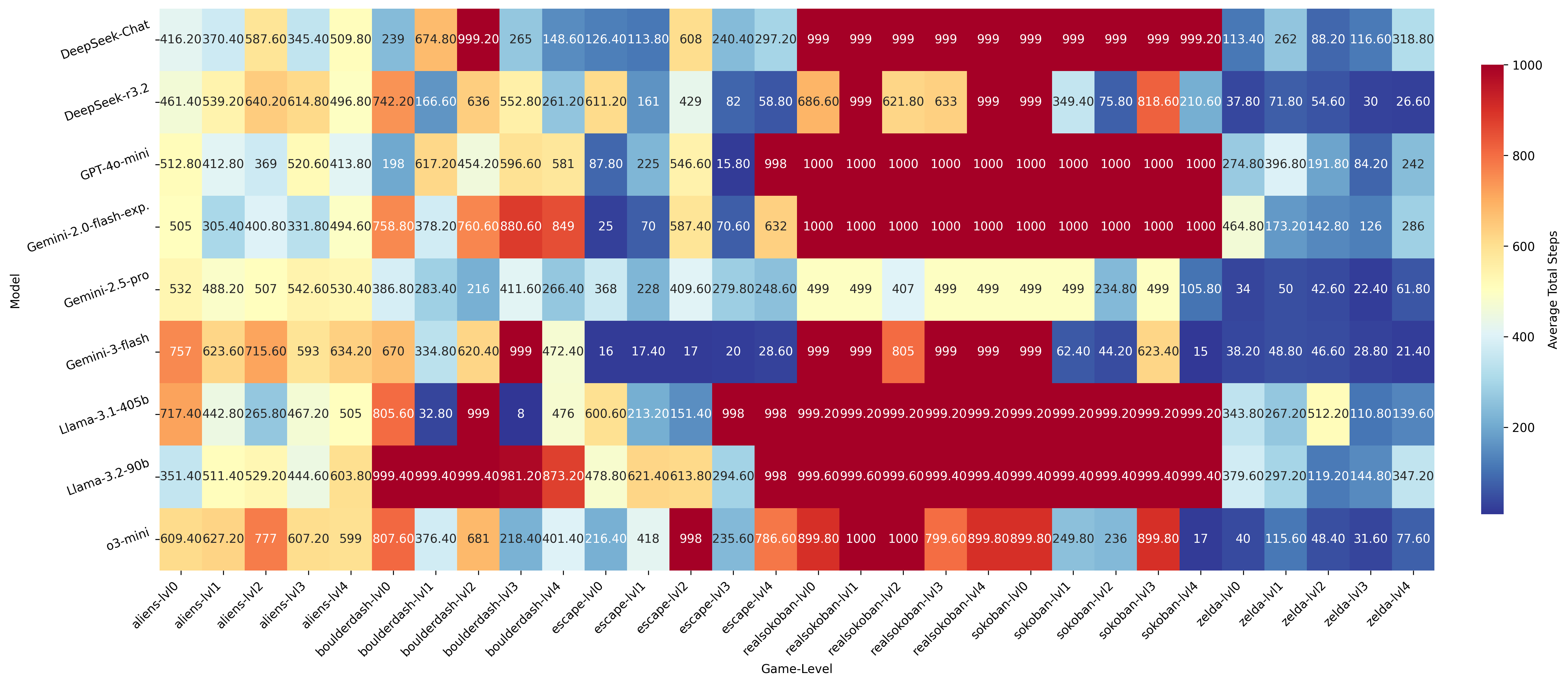}
  \caption{Zero-shot prompt: total steps heatmap.}
\end{figure*}

\clearpage
\subsection{Contextual Prompt}

\begin{figure*}[htbp]
  \centering
  \includegraphics[width=\textwidth]{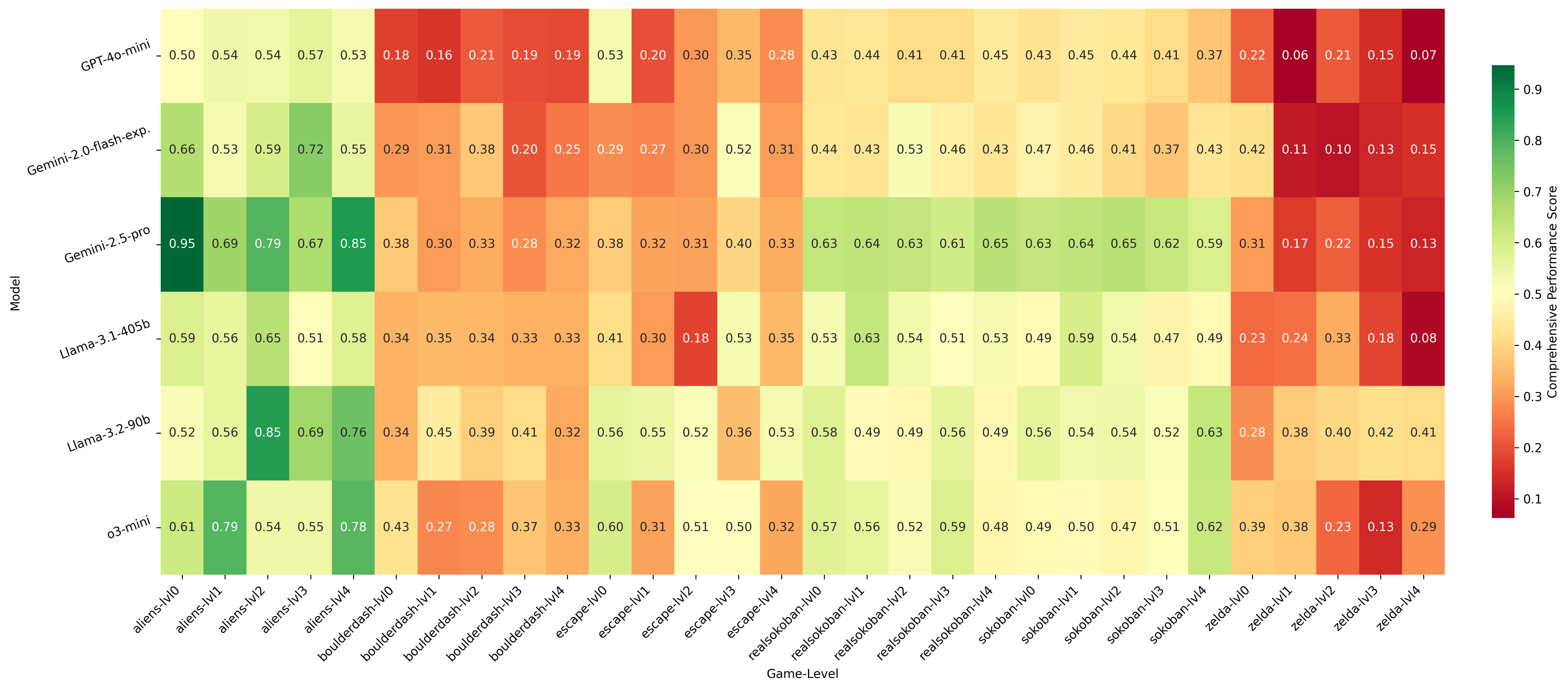}
  \caption{Contextual prompt: comprehensive performance heatmap.}
  \label{fig:app:contextual:comprehensive}
\end{figure*}

The following figures present evaluation results under the contextual prompt setting, where agents are additionally provided with a short history of recent actions and outcomes. The comprehensive performance heatmap (Figure~\ref{fig:app:contextual:comprehensive}) summarizes overall behavioral scores, followed by breakdowns of win rate, normalized reward, meaningful step ratio, and total steps. These results illustrate how access to recent interaction history affects agent performance across different game types.

\begin{figure*}[htbp]
  \centering
  \includegraphics[width=\textwidth]{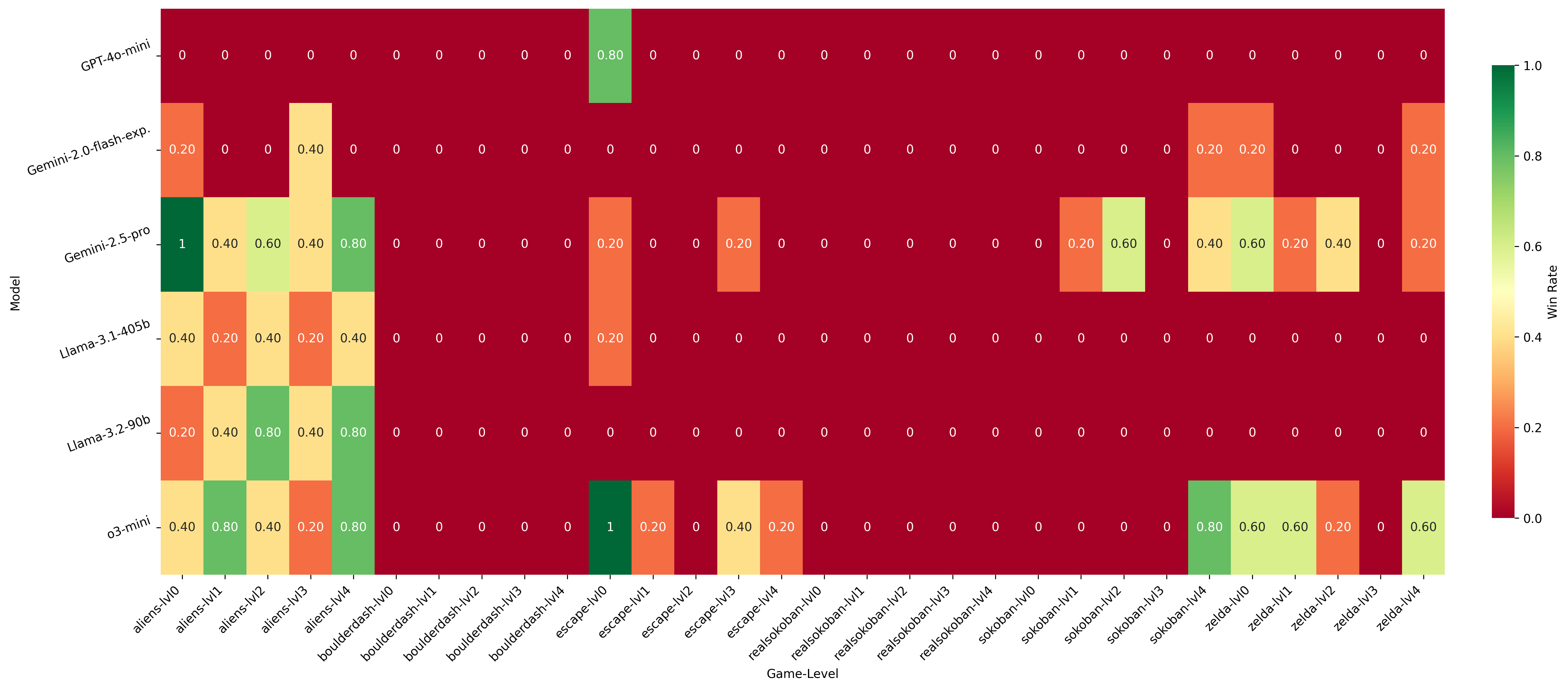}
  \caption{Contextual prompt: win rate heatmap.}
\end{figure*}

\begin{figure*}[htbp]
  \centering
  \includegraphics[width=\textwidth]{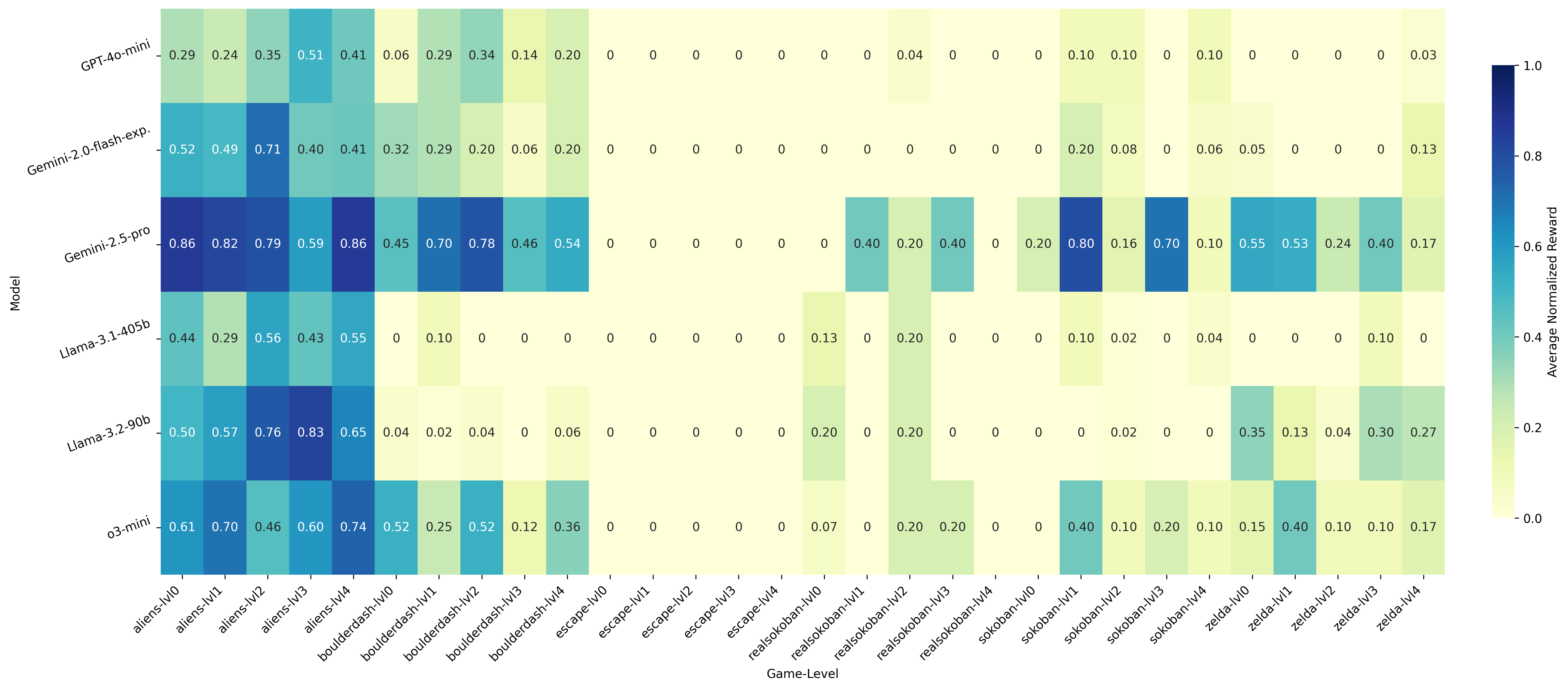}
  \caption{Contextual prompt: normalized reward heatmap.}
\end{figure*}

\begin{figure*}[htbp]
  \centering
  \includegraphics[width=\textwidth]{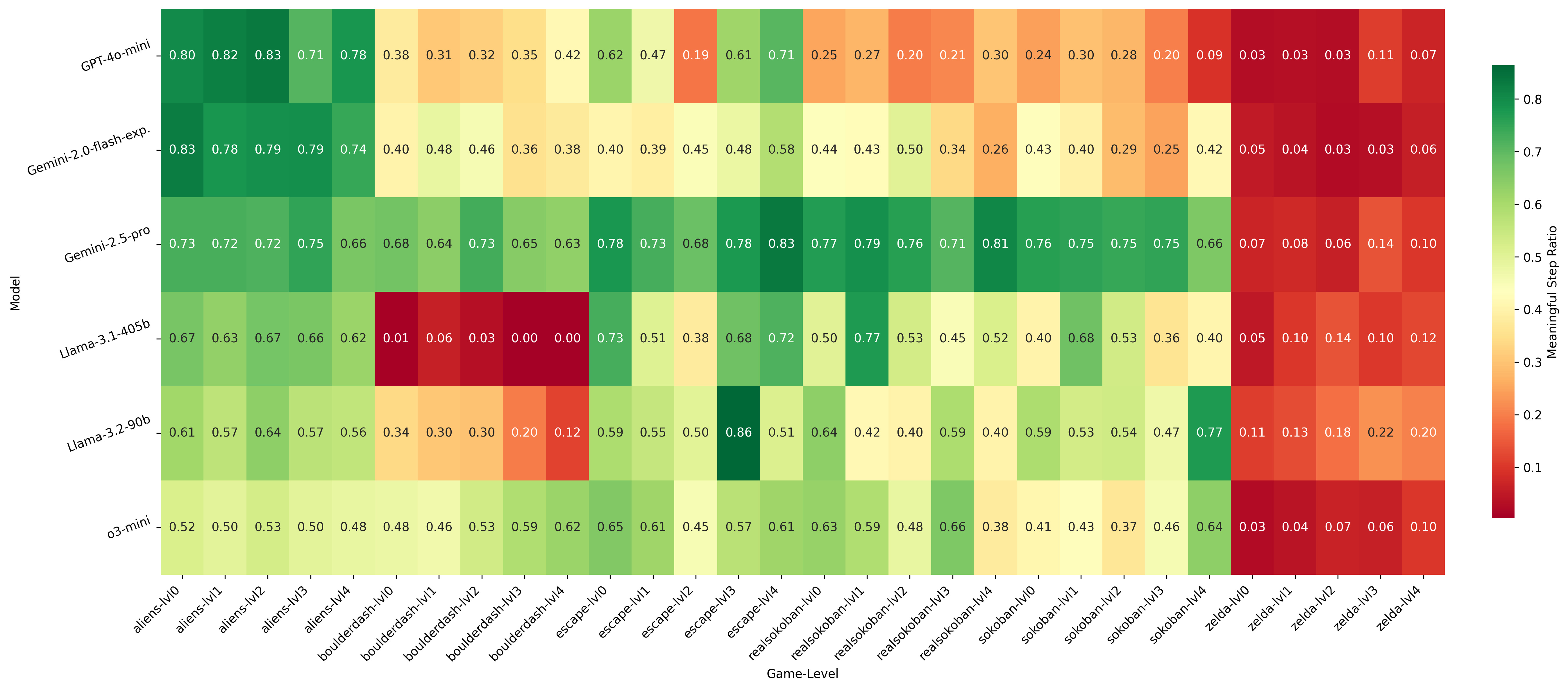}
  \caption{Contextual prompt: meaningful step ratio heatmap.}
\end{figure*}

\begin{figure*}[htbp]
  \centering
  \includegraphics[width=\textwidth]{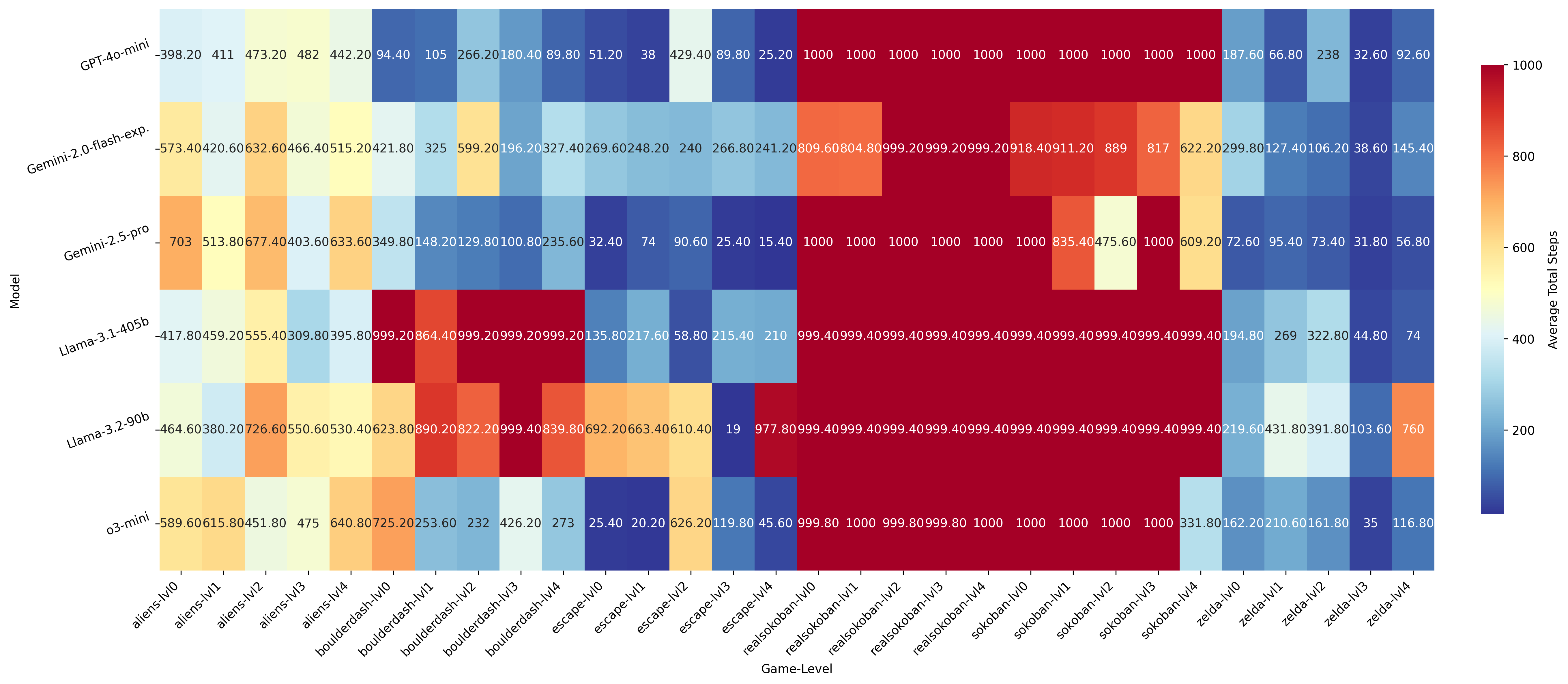}
  \caption{Contextual prompt: total steps heatmap.}
\end{figure*}

\clearpage
\newpage
\subsection{Both Prompt}

\begin{figure*}[htbp]
  \centering
  \includegraphics[width=\textwidth]{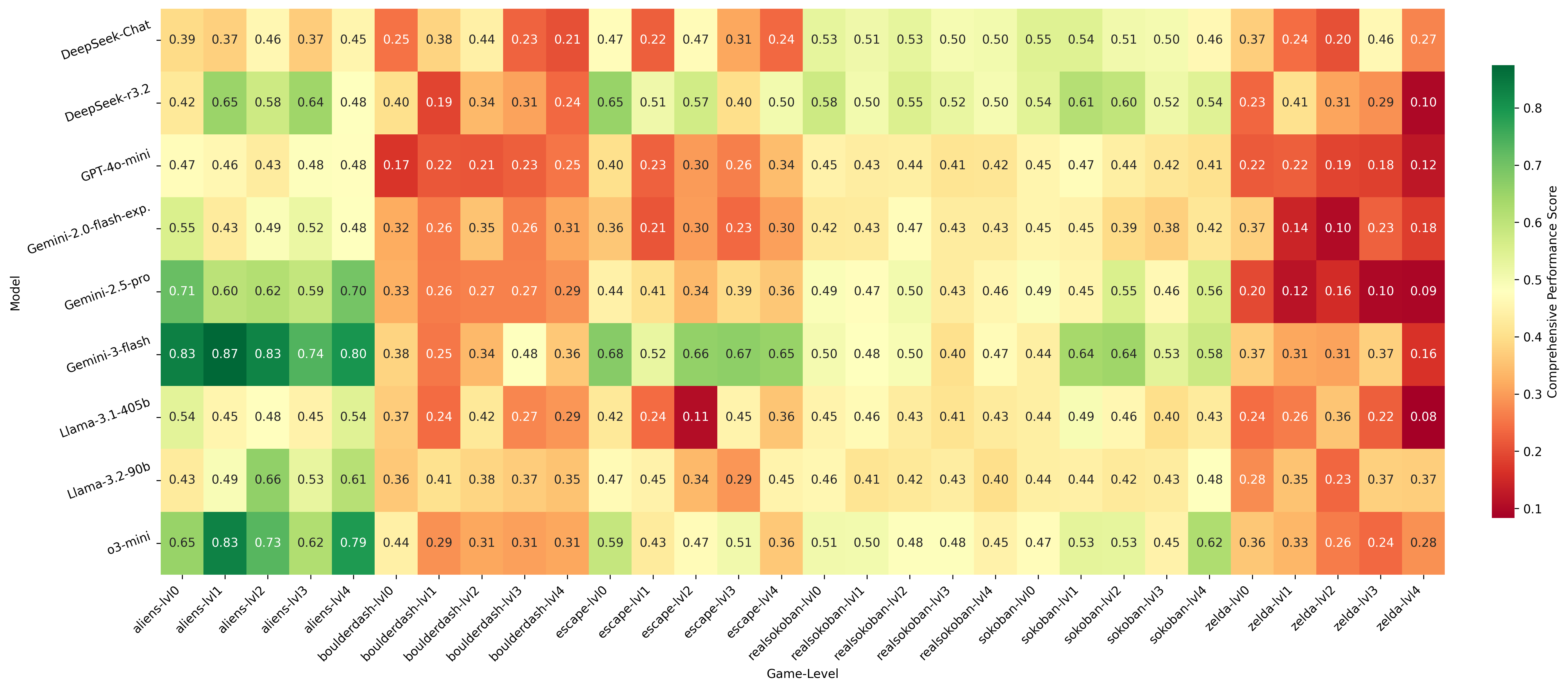}
  \caption{Overall prompt: comprehensive performance heatmap.}
  \label{fig:app:overall:comprehensive}
\end{figure*}

The following figures present aggregated evaluation results combining both the zero-shot and contextual prompt settings. The comprehensive performance heatmap (Figure~\ref{fig:app:overall:comprehensive}) provides an overall summary of agent behavioral scores, with subsequent heatmaps detailing win rate, normalized reward, meaningful step ratio, and total steps. This combined view offers a holistic perspective on agent capabilities across all evaluated prompt conditions.

\begin{figure*}[htbp]
  \centering
  \includegraphics[width=\textwidth]{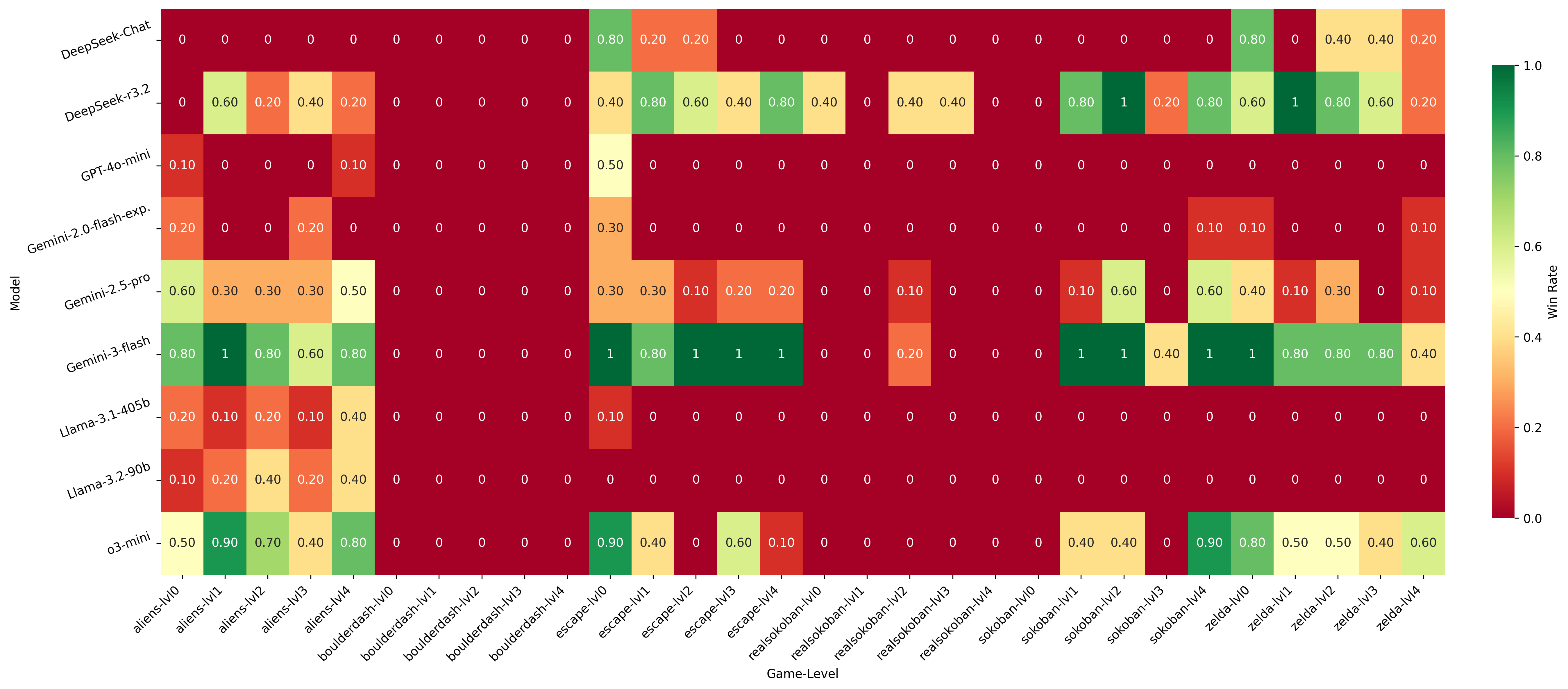}
  \caption{Overall prompt: win rate heatmap.}
\end{figure*}

\begin{figure*}[htbp]
  \centering
  \includegraphics[width=\textwidth]{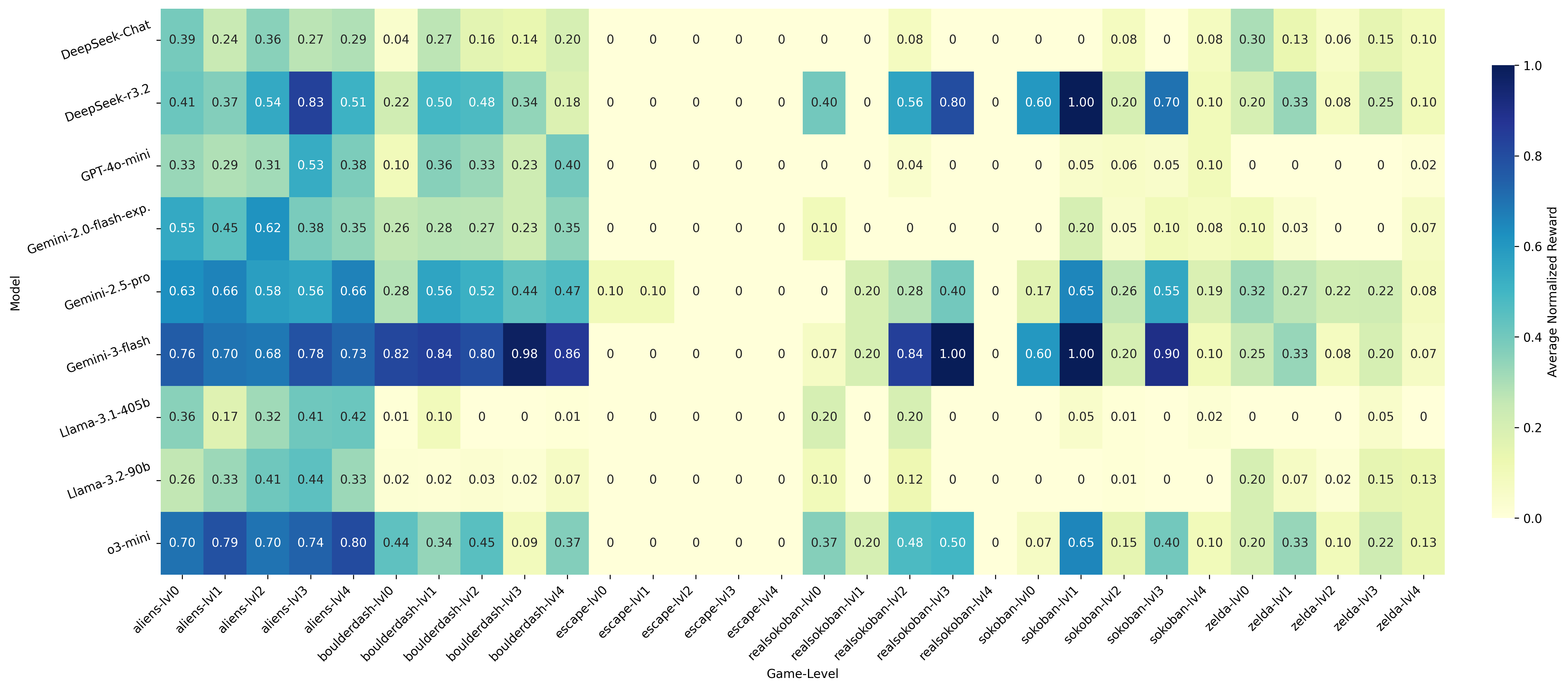}
  \caption{Overall prompt: normalized reward heatmap.}
\end{figure*}

\begin{figure*}[htbp]
  \centering
  \includegraphics[width=\textwidth]{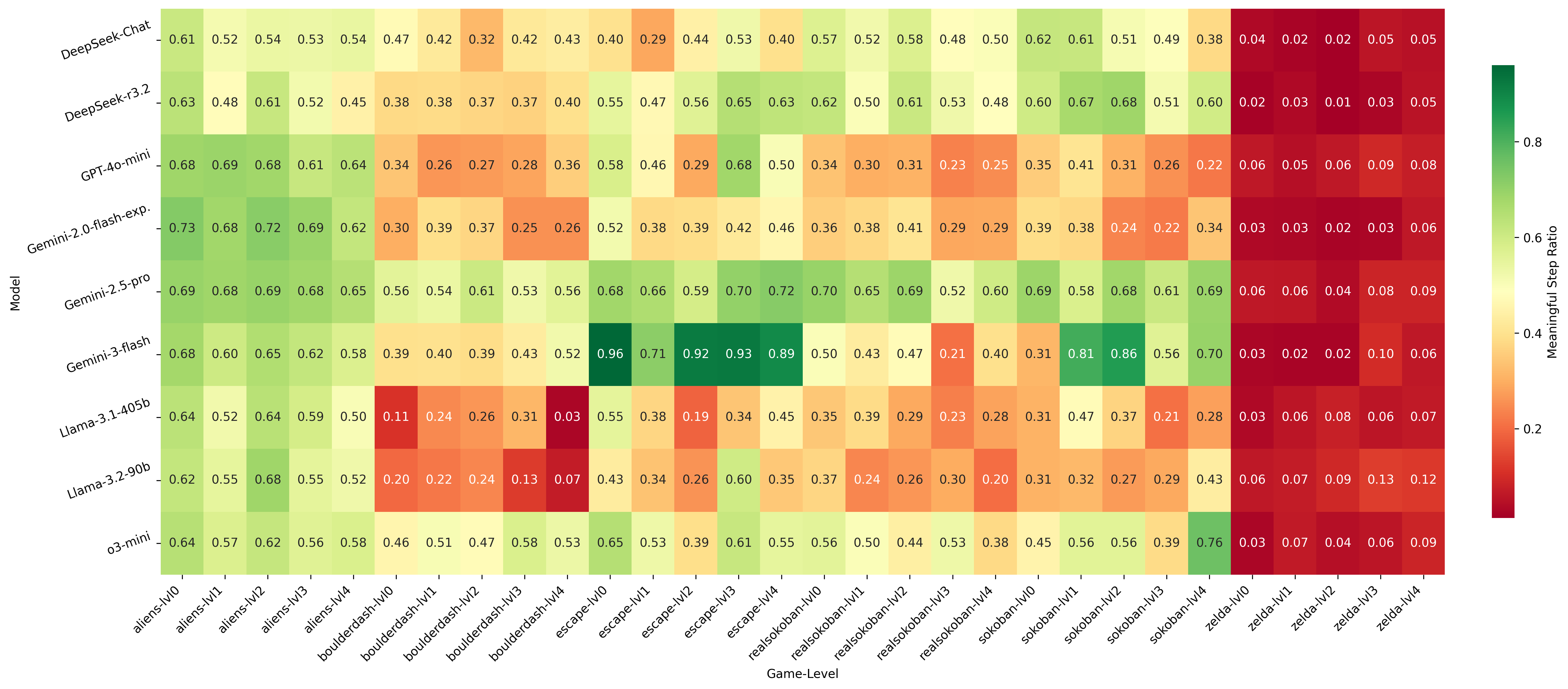}
  \caption{Overall prompt: meaningful step ratio heatmap.}
\end{figure*}

\begin{figure*}[htbp]
  \centering
  \includegraphics[width=\textwidth]{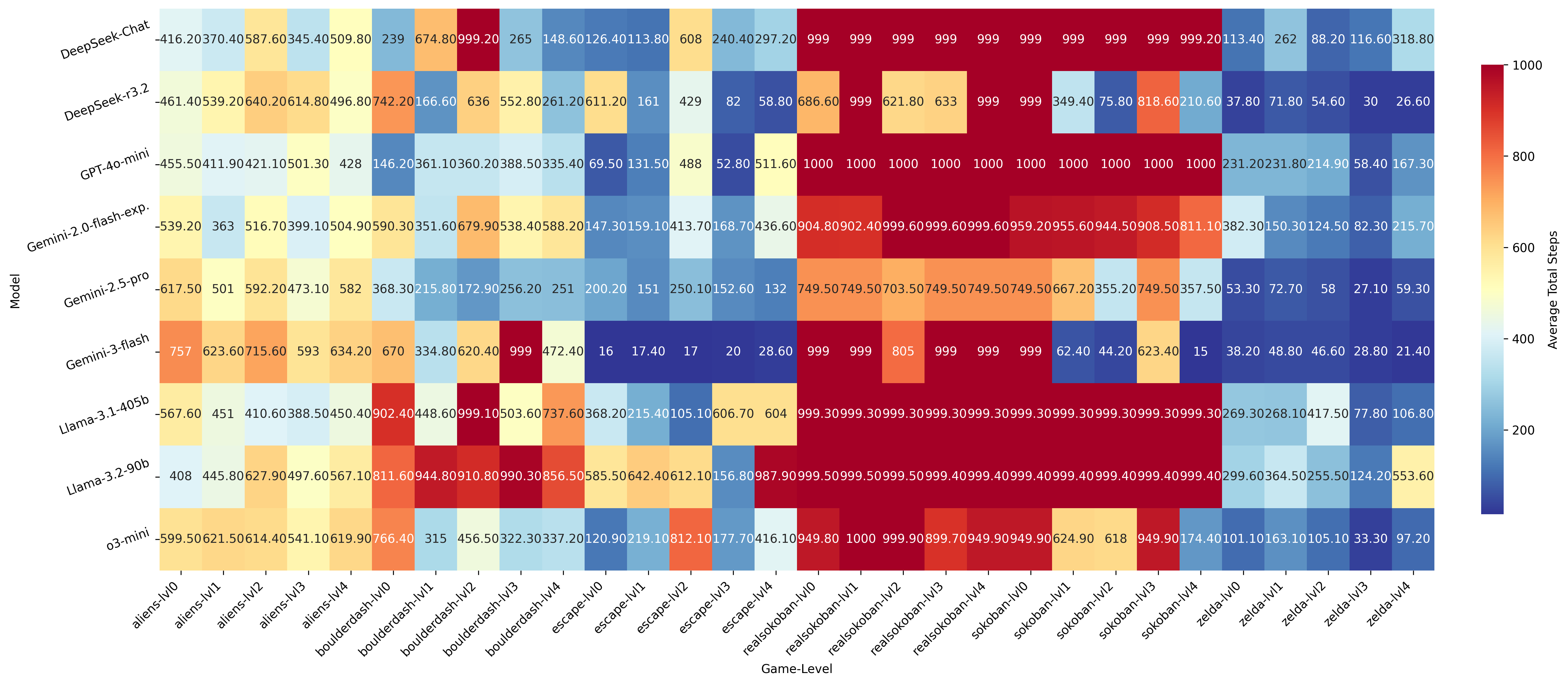}
  \caption{Overall prompt: total steps heatmap.}
\end{figure*}
\clearpage
\newpage
\newcommand{\wci}[3]{%
  \textbf{#1}{\,\tiny[#2,\,#3]}%
}
\newcommand{\hi}[3]{%
  \cellcolor{green!18}\textbf{#1}{\,\tiny[#2,\,#3]}%
}
\newcommand{\zz}[1]{%
  \textit{0.00}{\,\tiny[0.00,\,#1]}%
}


\section{Statistical Methodology}
\label{app:stats_methods}

The main paper reports aggregate performance metrics across 2,250 episodes:
all 9 models were evaluated under the zero-shot prompt ($9 \times 30
\text{ game-levels} \times 5 \text{ runs} = 1{,}350$ episodes), while 6 of the
9 models were additionally evaluated under the contextual prompt
($6 \times 30 \times 5 = 900$ episodes).
The three models evaluated only in zero-shot mode are DeepSeek-Chat,
DeepSeek-R3.2, and Gemini-3-flash.
This section describes the inferential methods applied post-hoc to those logs
without running any additional game episodes.  We address four questions:
(i) how reliable are the reported win rates given $n=5$ per cell?
(ii) does prompt strategy (zero-shot vs.\ contextual) significantly change
win rates for the six models where both were run?
(iii) are the four score components redundant, or do they capture independent
aspects of behaviour? and (iv) can the qualitative failure patterns described
in Section~5 be confirmed quantitatively from raw trajectory logs?

\paragraph{Wilson score confidence intervals.}
Win rates are reported as proportions $\hat{p} = k/n$ where $k$ is
the number of winning episodes out of $n$ trials.
Because standard Wald intervals are known to be unreliable at extreme
proportions (i.e.\ $\hat{p}=0$ or $1$), we use the Wilson score
interval~\cite{wilson1927probable} at the 95\% level ($z=1.96$):
\begin{equation}
  \tilde{p} \;\pm\; z
  \frac{\sqrt{\hat{p}(1-\hat{p})/n + z^{2}/(4n^2)}}{1+z^{2}/n},
  \qquad
  \tilde{p} = \frac{\hat{p}+z^{2}/(2n)}{1+z^{2}/n}.
\end{equation}

\paragraph{Fisher's exact test.}
We test two families of hypotheses using two-sided Fisher's exact
tests~\cite{fisher1935design}: (A) whether the prompt strategy
(zero-shot vs.\ contextual) changes the win rate for each
(model,~game-level) pair, and (B) whether any two models differ in win
rate on a given (game-level, prompt-type) cell.
With 55 (family A) and 697 (family B) simultaneous comparisons, raw
$p$-values are corrected for the false discovery rate using the
Benjamini--Hochberg (BH) procedure~\cite{benjamini1995controlling}.
Effect size is Cohen's $h = 2\arcsin\!\sqrt{p_1} - 2\arcsin\!\sqrt{p_2}$,
with $|h|= 0.2 / 0.5 / 0.8$ mapping to small/medium/large effects.

\paragraph{Score-component correlation.}
We compute Pearson and Spearman correlation matrices over all
$N=2{,}250$ individual episode records across the four reported
metrics: win/loss, meaningful-step ratio, total steps, and total
movement distance.  $p$-values are Bonferroni-corrected over the six
unique pairs.

\paragraph{Failure-mode metrics.}
The following trajectory statistics are computed per episode from the
raw \texttt{step\_metrics.json} logs:
\begin{itemize}[leftmargin=1.5em,itemsep=1pt]
  \item \emph{Bump rate}: fraction of steps in which the agent's
        grid position did not change (non-meaningful steps).
  \item \emph{Max.\ consecutive bumps}: longest unbroken run of
        non-meaningful steps—a proxy for being stuck.
  \item \emph{Late bump rate}: bump rate in the final 25\% of an
        episode, capturing terminal paralysis.
  \item \emph{Oscillation rate}: fraction of steps that immediately
        reverse the prior move ($A \!\to\! B \!\to\! A$ pattern).
  \item \emph{Unique cells}: number of distinct grid positions visited.
  \item \emph{Coverage density}: unique cells divided by bounding-box
        area—low values indicate repetitive pacing.
\end{itemize}
Differences between winning and losing episodes are assessed with
Welch's two-sample $t$-tests (no multiple-comparison correction).

\section{Win Rates with 95\% Confidence Intervals}
\label{app:wilson}

With exactly $n=5$ episodes per game-level for every model, point estimates
at the extremes of 0\% and 100\% are particularly sensitive to sample
variability.  Table~\ref{tab:winrate_ci} reports zero-shot win rates
aggregated over all five levels of each game (total $n=25$ per cell) together
with 95\% Wilson confidence intervals, which remain well-calibrated even at
extreme proportions.

Two broad patterns emerge.  First, BoulderDash yields 0\% wins for every
model and every level; its CIs bound plausible performance from above at
$[0.00, 0.13]$.
Second, three frontier models—Gemini-3-flash, o3-mini, and DeepSeek-R3.2—achieve
aggregated win rates $\ge 0.5$ on multiple games with lower bounds that
substantially exceed chance, whereas the remaining six models fail to surpass
0.5 on any game after CI adjustment.

\begin{figure}[htbp]
  \centering
  \includegraphics[width=0.85\textwidth]{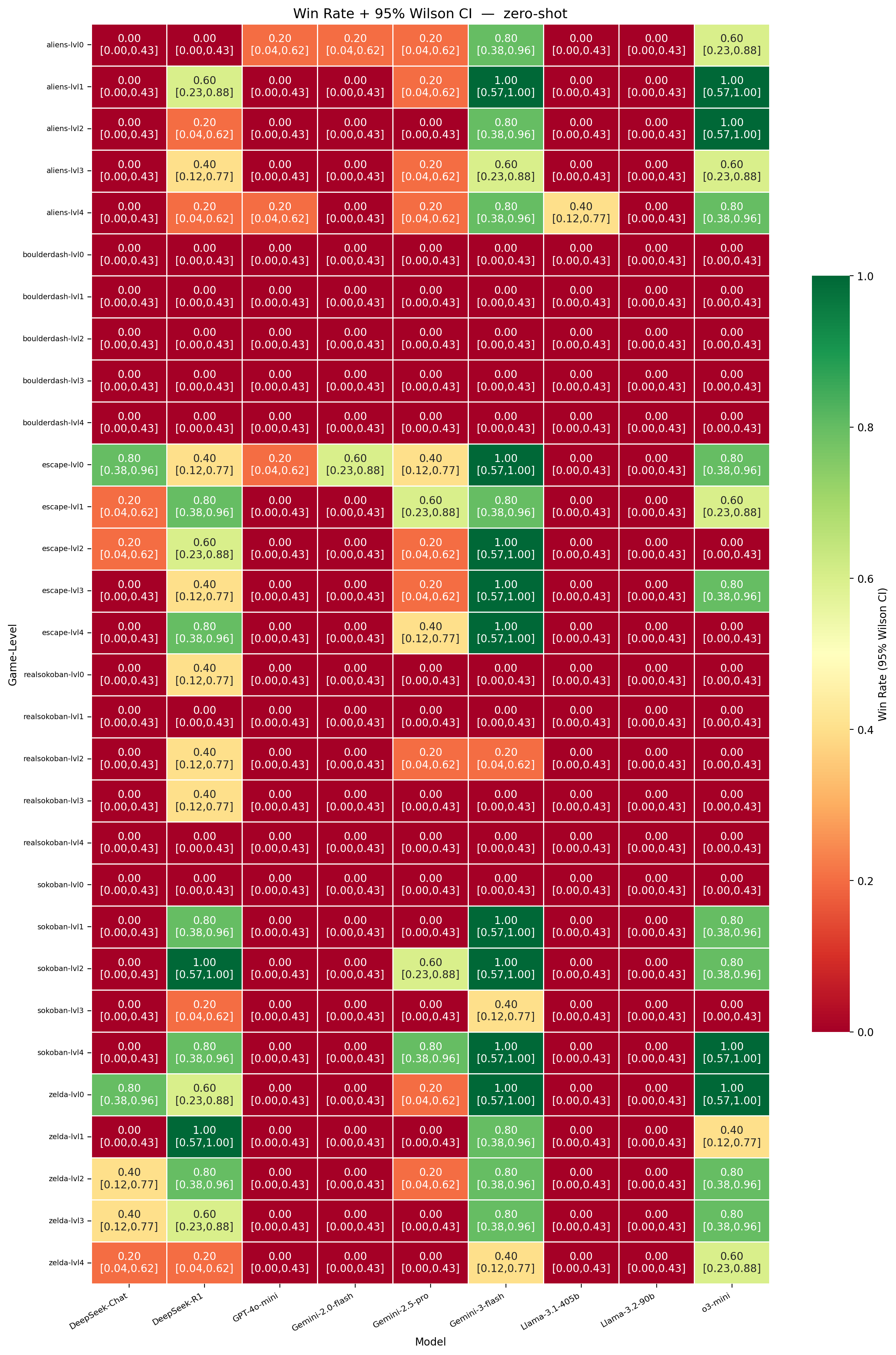}
  \caption{Win rate heatmap (zero-shot) with per-cell 95\% Wilson
  confidence intervals ($n=5$ per cell). Rows are game-levels;
  columns are models. BoulderDash levels are uniformly dark (0\%
  win rate across all models).}
  \label{fig:wilson_heatmap}
\end{figure}

\begin{figure}[htbp]
  \centering
  \includegraphics[width=\textwidth]{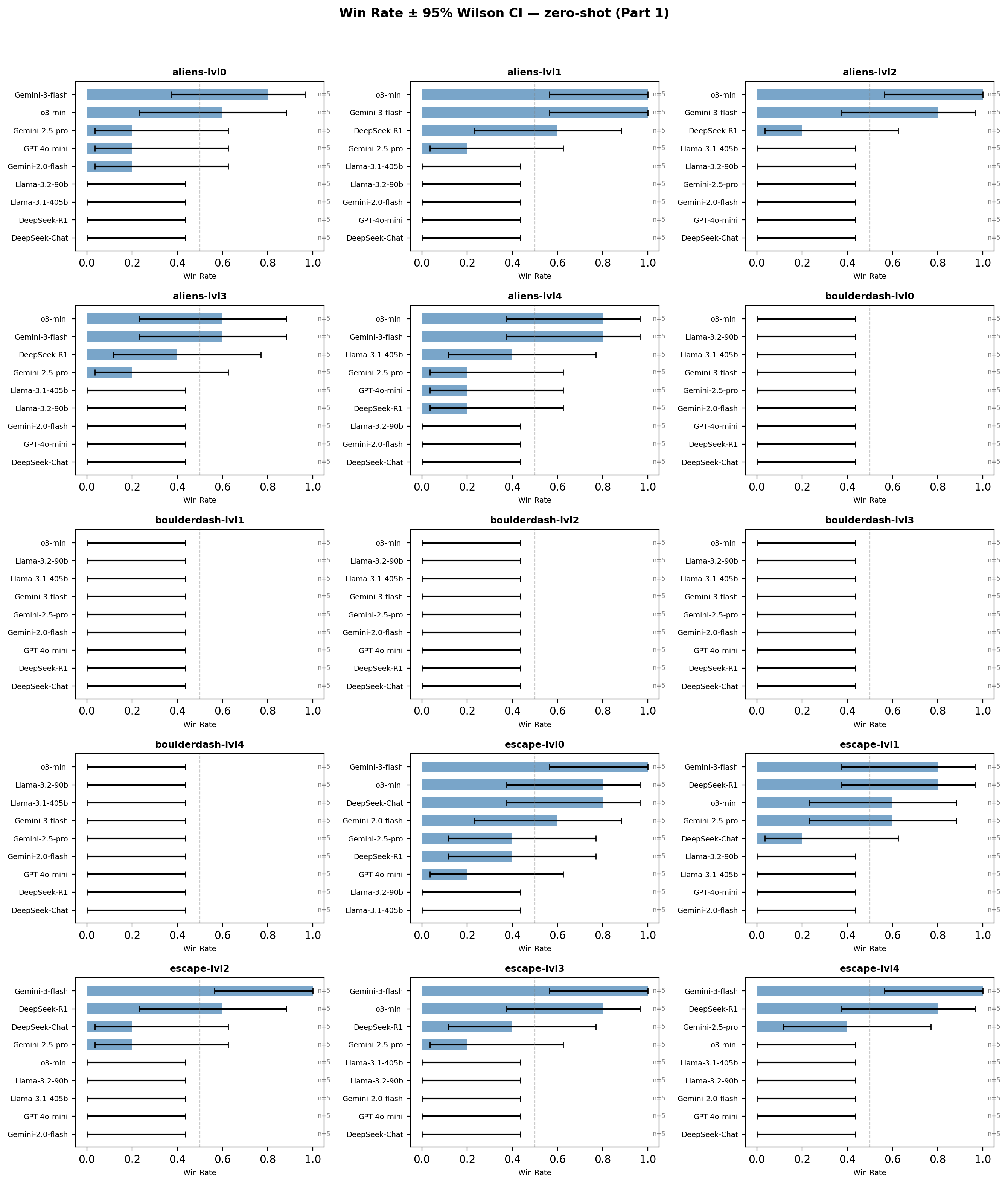}
  \caption{Forest plots of zero-shot win rates ($\pm$95\% Wilson CI),
  Part~1: Aliens and BoulderDash (levels 0--4 each).
  Models sorted by point estimate within each panel;
  dashed line marks $p=0.5$.}
  \label{fig:wilson_forest1}
\end{figure}

\begin{figure}[htbp]
  \centering
  \includegraphics[width=\textwidth]{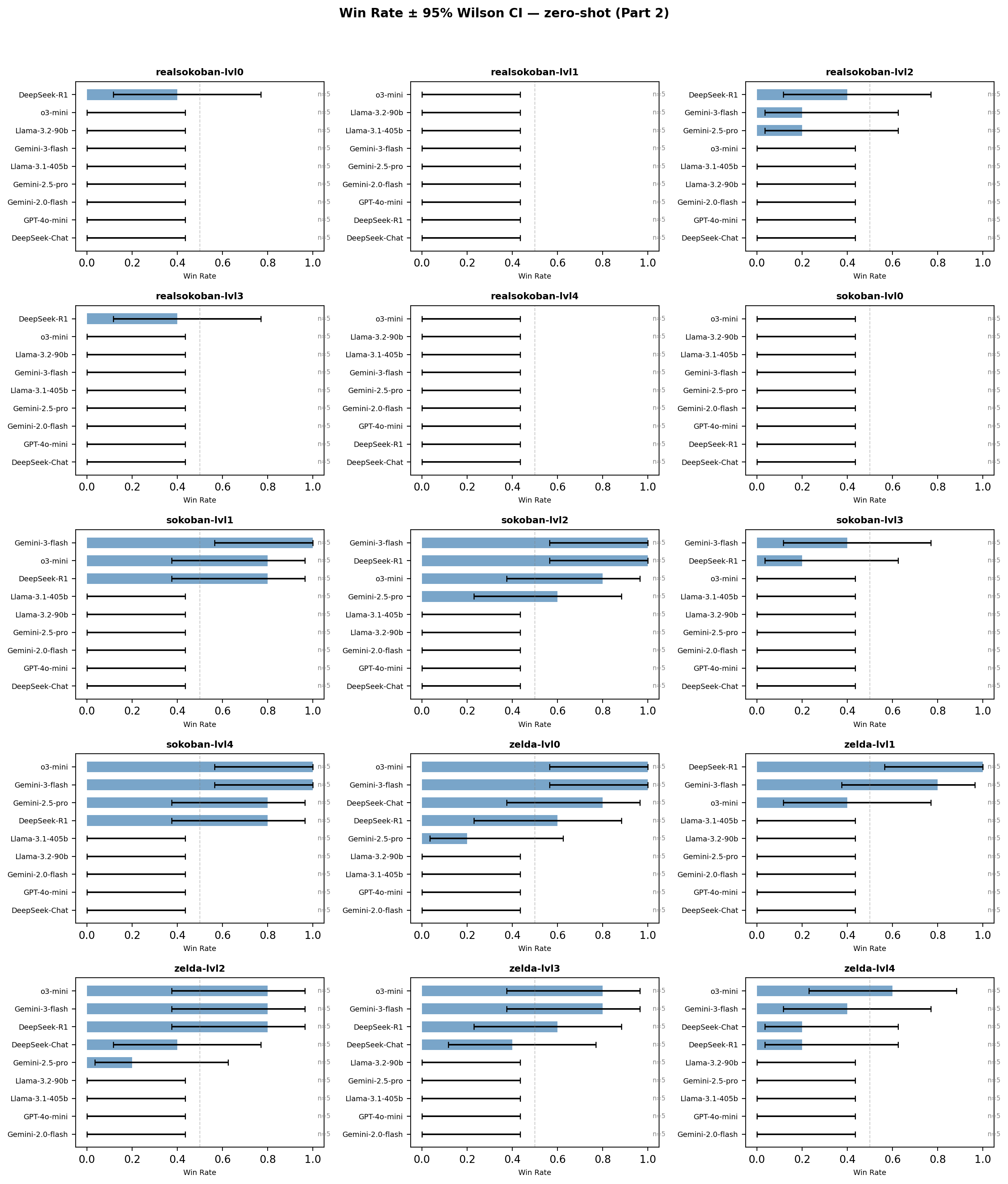}
  \caption{Forest plots of zero-shot win rates ($\pm$95\% Wilson CI),
  Part~2: Escape, RealSokoban, Sokoban, and Zelda (levels 0--4 each).}
  \label{fig:wilson_forest2}
\end{figure}

Figure~\ref{fig:wilson_heatmap} shows the heatmap overview;
Figures~\ref{fig:wilson_forest1} and~\ref{fig:wilson_forest2} provide
the per-level forest-plot breakdown split across the six games.

\begin{table}[htbp]
\centering
\caption{Zero-shot win rate (aggregated over levels 0--4, $n=25$ per cell)
with 95\% Wilson confidence intervals. Format: \textbf{rate}~{\tiny[lo,\,hi]}.
Cells with rate $\ge0.5$ are shaded green; zero-win cells are italicised.
BD = BoulderDash, RS = RealSokoban.}
\label{tab:winrate_ci}
\setlength{\tabcolsep}{4pt}
\small
\begin{tabular}{lcccccc}
\toprule
\textbf{Model} & \textbf{Aliens} & \textbf{BD} & \textbf{Escape}
               & \textbf{RS} & \textbf{Sokoban} & \textbf{Zelda} \\
\midrule
DeepSeek-Chat
  & \zz{.13}
  & \zz{.13}
  & \wci{0.24}{.11}{.43}
  & \zz{.13}
  & \zz{.13}
  & \wci{0.36}{.20}{.55} \\
DeepSeek-R3.2
  & \wci{0.28}{.14}{.48}
  & \zz{.13}
  & \hi{0.60}{.41}{.77}
  & \wci{0.24}{.11}{.43}
  & \hi{0.56}{.37}{.73}
  & \hi{0.64}{.45}{.80} \\
GPT-4o-mini
  & \wci{0.08}{.02}{.25}
  & \zz{.13}
  & \wci{0.04}{.01}{.20}
  & \zz{.13}
  & \zz{.13}
  & \zz{.13} \\
Gemini-2.0-flash
  & \wci{0.04}{.01}{.20}
  & \zz{.13}
  & \wci{0.12}{.04}{.30}
  & \zz{.13}
  & \zz{.13}
  & \zz{.13} \\
Gemini-2.5-pro
  & \wci{0.16}{.06}{.35}
  & \zz{.13}
  & \wci{0.36}{.20}{.55}
  & \wci{0.04}{.01}{.20}
  & \wci{0.28}{.14}{.48}
  & \wci{0.08}{.02}{.25} \\
Gemini-3-flash
  & \hi{0.80}{.61}{.91}
  & \zz{.13}
  & \hi{0.96}{.80}{.99}
  & \wci{0.04}{.01}{.20}
  & \hi{0.68}{.48}{.83}
  & \hi{0.76}{.57}{.89} \\
Llama-3.1-405b
  & \wci{0.08}{.02}{.25}
  & \zz{.13}
  & \zz{.13}
  & \zz{.13}
  & \zz{.13}
  & \zz{.13} \\
Llama-3.2-90b
  & \zz{.13}
  & \zz{.13}
  & \zz{.13}
  & \zz{.13}
  & \zz{.13}
  & \zz{.13} \\
o3-mini
  & \hi{0.80}{.61}{.91}
  & \zz{.13}
  & \wci{0.44}{.27}{.63}
  & \zz{.13}
  & \hi{0.52}{.33}{.70}
  & \hi{0.72}{.52}{.86} \\
\bottomrule
\end{tabular}
\end{table}

\section{Statistical Significance of Performance Differences}
\label{app:fisher}

Fisher's exact test is used throughout because many cells contain counts
of 5--10, making asymptotic $\chi^2$ approximations unreliable.  Two
families of tests are reported.

\subsection{Prompt Strategy Effect (Zero-Shot vs.\ Contextual)}

A recurring question in LLM benchmark design is whether providing the agent
with recent interaction history (contextual prompting) materially changes
performance relative to a pure zero-shot setting.
We tested all 55 (model, game-level) pairs where both prompt strategies were
evaluated. After BH-FDR correction, \textbf{zero} comparisons reach
significance at $\alpha=0.05$.  We therefore cannot reject the null hypothesis
that contextual prompting changes win rates; any apparent differences in the
main paper's heatmaps should be treated as noise at the sample sizes used.
This null result suggests the bottleneck is spatial reasoning rather than
memory of recent actions.

\subsection{Pairwise Model Comparisons}

Across 697 two-sided Fisher tests (all model pairs within each
game-level $\times$ prompt-type cell; zero-shot includes all 9 models,
contextual includes the 6 models with contextual data),
\textbf{zero pairs} are significant at $p_\text{adj}<0.05$ after
BH-FDR correction.  This is a power limitation inherent to $n=5$ per cell:
even a perfect separation (5/5 vs.\ 0/5) yields an uncorrected
$p\approx0.008$, which inflates to $p_\text{adj}>1$ after BH adjustment
across 697 tests—well above any reasonable threshold.

Although formal significance cannot be claimed, the \emph{effect sizes} are
substantial.  Table~\ref{tab:fisher_selected} lists a representative set of
high-contrast pairs with Cohen's $h\ge2.5$ (near-complete win-rate
separation), grouped by game-level.  These gaps are practically meaningful
regardless of the power limitation, and are corroborated by the Wilson CIs
in Section~\ref{app:wilson}, where the intervals for top-performing and
bottom-performing models on the same game do not overlap.

\begin{table}[htbp]
\centering
\caption{Representative high-contrast pairwise win-rate comparisons
(zero-shot; $n=5$ per model per level).
$h$ = Cohen's $h$ (effect size); $|h|=0.8$ is ``large''.
No pairs reach BH-adjusted significance at $n=5$ (see text);
rows show uncorrected $p$ as indicative effect-size evidence.}
\label{tab:fisher_selected}
\setlength{\tabcolsep}{4pt}
\small
\begin{tabular}{llcccccr}
\toprule
\textbf{Model A} & \textbf{Model B} & \textbf{Game-Level}
  & $r_A$ & $r_B$ & $p_\text{raw}$ & $|h|$ & \\
\midrule
o3-mini          & Llama-3.1-405b & Zelda-lvl0   & 1.00 & 0.00 & .008 & 3.14 \\
o3-mini          & Llama-3.2-90b  & Zelda-lvl0   & 1.00 & 0.00 & .008 & 3.14 \\
Gemini-3-flash   & Llama-3.2-90b  & Zelda-lvl0   & 1.00 & 0.00 & .008 & 3.14 \\
\midrule
DeepSeek-R3.2      & GPT-4o-mini    & Sokoban-lvl2 & 1.00 & 0.00 & .008 & 3.14 \\
Gemini-3-flash   & Llama-3.2-90b  & Sokoban-lvl4 & 1.00 & 0.00 & .008 & 3.14 \\
\midrule
Gemini-3-flash   & GPT-4o-mini    & Aliens-lvl1  & 1.00 & 0.00 & .008 & 3.14 \\
o3-mini          & Llama-3.2-90b  & Aliens-lvl1  & 1.00 & 0.00 & .008 & 3.14 \\
\midrule
DeepSeek-R3.2      & Llama-3.2-90b  & Escape-lvl1  & 1.00 & 0.00 & .008 & 3.14 \\
Gemini-3-flash   & GPT-4o-mini    & Escape-lvl0  & 1.00 & 0.00 & .008 & 3.14 \\
\bottomrule
\multicolumn{8}{l}{$p_\text{raw}$: uncorrected Fisher's exact (two-sided);
  none survive BH-FDR at $n=5$.  $h=3.14$ = 100\% vs.\ 0\%.}
\end{tabular}
\end{table}

\section{Score Component Correlation Analysis}
\label{app:correlation}

The benchmark reports four episode-level statistics: binary win/loss
(\textbf{Win}), meaningful-step ratio (\textbf{MSR}), total steps
taken (\textbf{Steps}), and total Manhattan displacement
(\textbf{Dist}).  A key question is whether these four numbers are
largely redundant—if they correlate at $|r|>0.8$, the ``comprehensive
score'' in the main paper adds no information beyond any single metric.

\begin{figure}[htbp]
  \centering
  \includegraphics[width=0.55\textwidth]{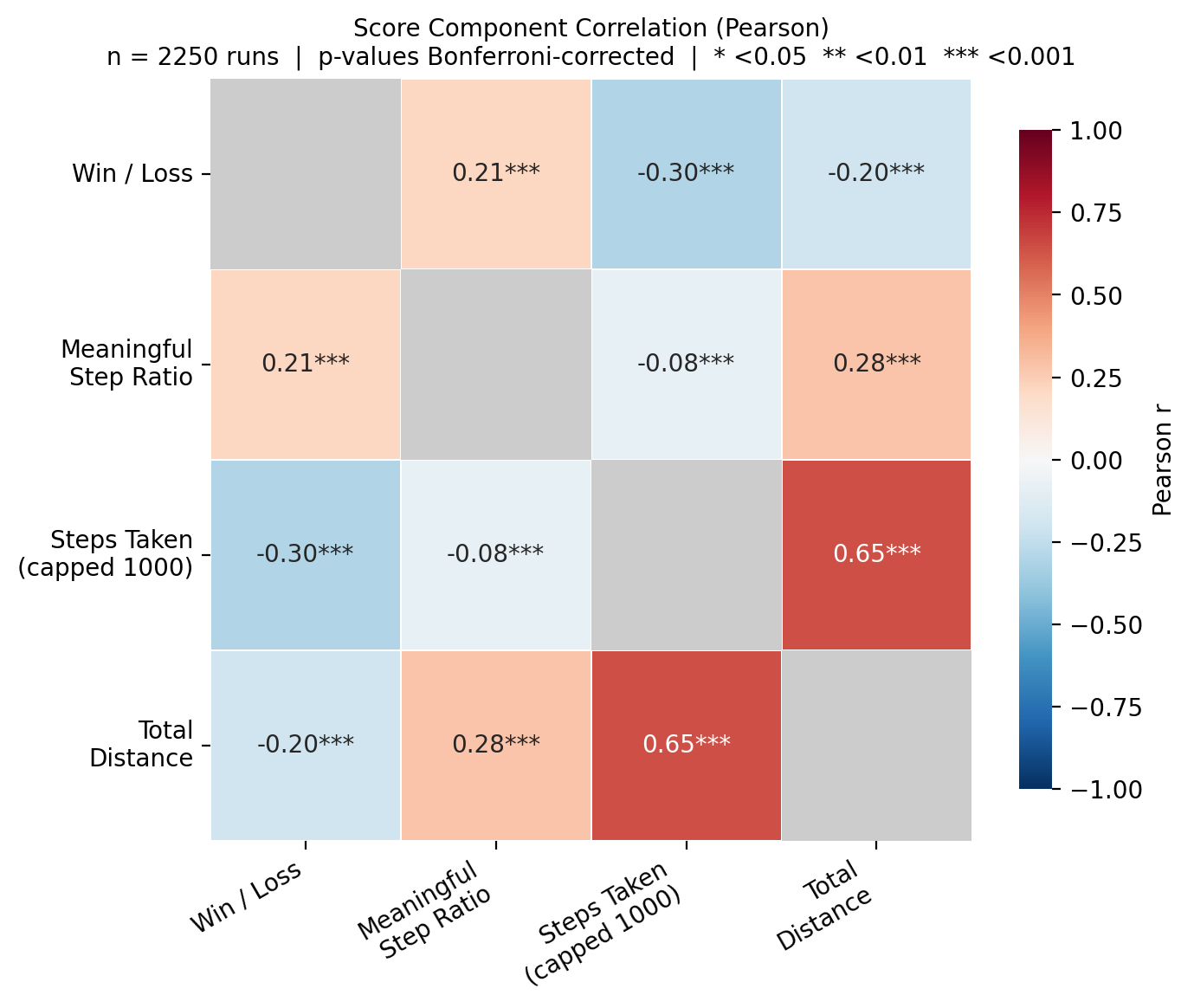}
  \caption{Pearson correlation matrix of score components
  ($N=2{,}250$ episodes). Significance markers follow Table~\ref{tab:correlation}.
  No pair exceeds $|r|=0.7$, confirming that all four metrics
  capture non-redundant information.}
  \label{fig:correlation}
\end{figure}

Table~\ref{tab:correlation} shows Pearson correlations over all
$N=2{,}250$ episode records, with Bonferroni-corrected $p$-values.
Figure~\ref{fig:correlation} visualises the same matrix.

\begin{table}[htbp]
\centering
\caption{Pearson correlation matrix of the four score components
($N = 2{,}250$ episodes). $p$-values are Bonferroni-corrected over 6
pairs. Significance: ${}^{*}p<.05$, ${}^{**}p<.01$,
${}^{***}p<.001$.}
\label{tab:correlation}
\setlength{\tabcolsep}{6pt}
\begin{tabular}{lcccc}
\toprule
 & \textbf{Win} & \textbf{MSR} & \textbf{Steps} & \textbf{Dist} \\
\midrule
\textbf{Win (binary)}
  & 1.00
  & $+0.21^{***}$
  & $-0.28^{***}$
  & $-0.18^{***}$ \\
\textbf{MSR}
  & $+0.21^{***}$
  & 1.00
  & $-0.10^{***}$
  & $+0.30^{***}$ \\
\textbf{Steps taken}
  & $-0.28^{***}$
  & $-0.10^{***}$
  & 1.00
  & $+0.62^{***}$ \\
\textbf{Total distance}
  & $-0.18^{***}$
  & $+0.30^{***}$
  & $+0.62^{***}$
  & 1.00 \\
\bottomrule
\end{tabular}
\end{table}

The largest correlation is between Steps and Dist ($r=0.62$): longer
episodes naturally cover more ground, so these two metrics are
moderately redundant.  Crucially, no pair exceeds $|r|=0.7$, so all
four metrics contribute \emph{non-redundant} variance to the
comprehensive score.  Win/loss correlates only modestly with the
continuous metrics ($|r| \le 0.28$), indicating that an agent can
score well on MSR or Dist while still losing—for example, an agent
that bumps walls repeatedly may exhaust the step budget (high Steps)
without ever reaching the goal.  The four components therefore capture
genuinely distinct facets of behaviour, justifying their combined use
in the main paper's scoring formula.

\section{Quantitative Failure Mode Analysis}
\label{app:failure}

Section~5 of the main paper discusses qualitative failure patterns
observed in agent trajectories.  Here we quantify these patterns across
$N=2{,}246$ of the 2{,}250 episodes (4 runs on \texttt{escape-lvl2}
produced empty step logs and are excluded), using six metrics extracted
from the step-level logs.

\subsection{Winning vs.\ Losing Episodes}

Table~\ref{tab:failure_modes} compares the six metrics between
winning ($n=347$) and losing ($n=1{,}899$) episodes.  Five of the six
metrics differ significantly between outcomes; only coverage density
does not.

\begin{table}[htbp]
\centering
\caption{Failure-mode metrics for winning ($n=347$) vs.\ losing ($n=1{,}899$)
episodes (means; Welch's $t$-test, no multiple-comparison correction).}
\label{tab:failure_modes}
\setlength{\tabcolsep}{5pt}
\begin{tabular}{lrrrrl}
\toprule
\textbf{Metric}
  & \textbf{Won} & \textbf{Lost}
  & \textbf{Diff.}
  & \textbf{$t$}  & \textbf{$p$} \\
\midrule
Bump rate
  & 0.464 & 0.624 & $-0.160$ & $-9.08$ & $<.001^{***}$ \\
Max.\ consec.\ bumps (steps)
  & 14.1  & 136.5 & $-122.4$ & $-14.76$ & $<.001^{***}$ \\
Late bump rate (last 25\%)
  & 0.467 & 0.627 & $-0.160$ & $-8.47$ & $<.001^{***}$ \\
Oscillation rate
  & 0.181 & 0.151 & $+0.030$ & $+2.83$  & $.005^{**}$ \\
Unique cells visited
  & 22.0  & 19.2  & $+2.8$   & $+4.63$  & $<.001^{***}$ \\
Coverage density
  & 0.688 & 0.711 & $-0.023$ & $-1.19$  & $.237^{\phantom{*}\text{ns}}$ \\
\bottomrule
\multicolumn{6}{l}{Significance: *** $p<.001$, ** $p<.01$, ns = not significant.}
\end{tabular}
\end{table}

The dominant discriminator between outcomes is
\textbf{max.\ consecutive bumps}: losing episodes average $137$ steps
of unbroken wall-bumping against only $14$ for winning ones
($|t|=14.8$, $p<.001$).  This metric captures the most severe failure
mode—an agent that enters a wall-bumping loop cannot recover, and the
episode is effectively over before the step budget expires.
Late bump rate mirrors this: agents that eventually lose enter stuck
patterns toward the end of the episode, not merely transiently.

Notably, \textbf{oscillation rate} ($A{\to}B{\to}A$ reversals) is
significantly \emph{higher} in winning episodes ($0.181$ vs.\ $0.151$,
$t=+2.83$, $p=.005$).  This reversal of the expected direction indicates
that oscillation is a signature of \emph{active exploration} rather than
mere confusion: agents that eventually succeed back-track more, suggesting
they are revising failed plans rather than perseverating.  Coverage density
($p=.237$) is not significantly different, confirming that the number of
cells visited per bounding-box area is a poor discriminator of success.

\subsection{Game-Specific Patterns}

\begin{figure}[htbp]
  \centering
  \includegraphics[width=\textwidth]{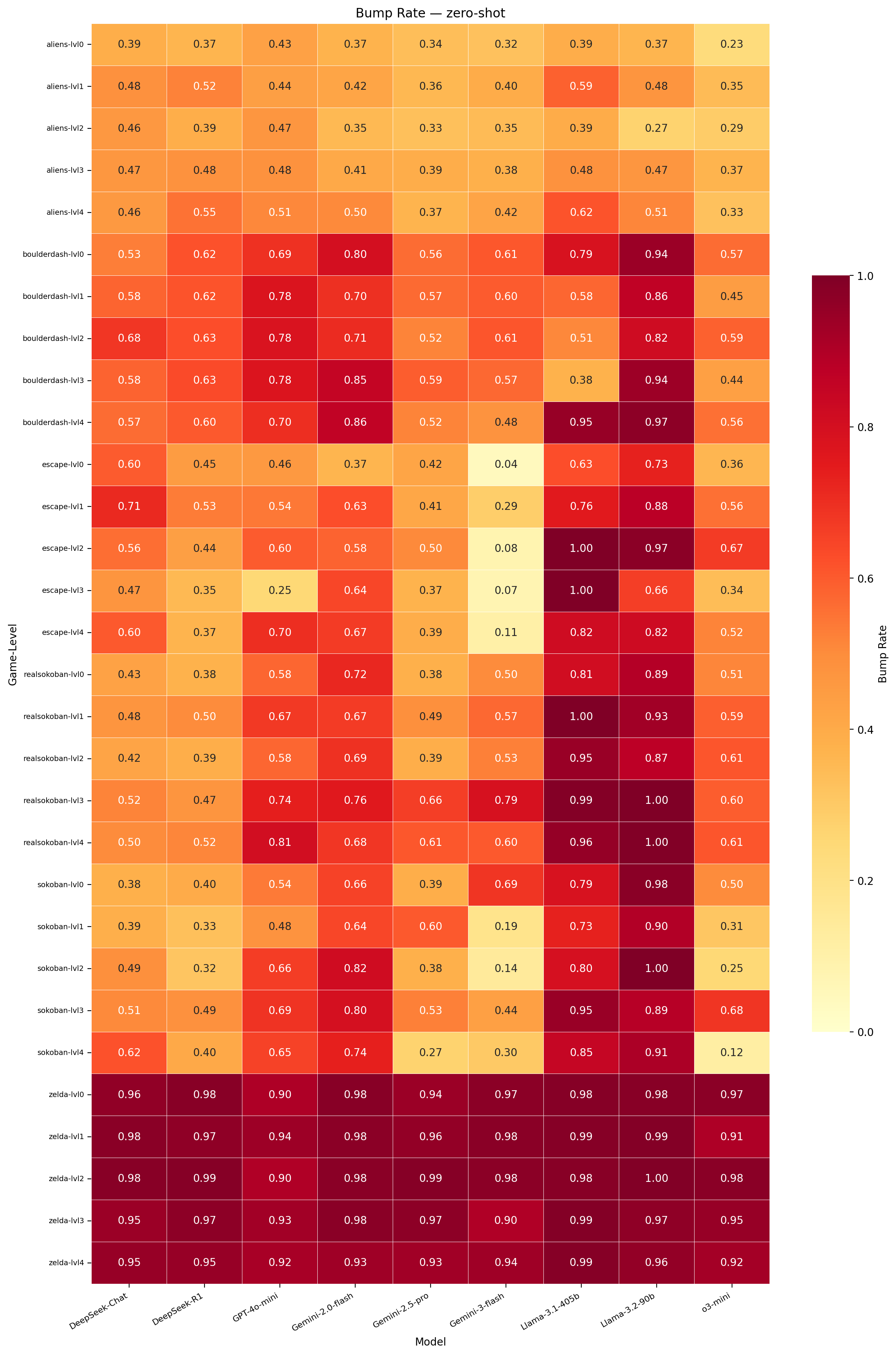}
  \caption{Mean bump rate (fraction of non-meaningful steps) by model
  and game-level (zero-shot). Higher values indicate more wasted
  actions. BoulderDash rows are uniformly dark ($>0.63$), reflecting
  the complete absence of any successful strategy across all models.}
  \label{fig:bump_heatmap}
\end{figure}

Figure~\ref{fig:bump_heatmap} shows the mean bump rate per
(model, game-level) in zero-shot mode.  Three patterns stand out.

\begin{itemize}[leftmargin=1.5em,itemsep=2pt]
  \item \textbf{BoulderDash}: all models exhibit bump rates of
        $0.63$--$0.69$, the highest across all games, consistent with
        the 0\% win rate.  The dynamic environment—rocks fall and crush
        the avatar—means that agents cannot formulate valid multi-step
        move sequences; most actions either collide with moving objects
        or are blocked by walls before any progress is made.
  \item \textbf{RealSokoban}: the top-10 oscillation-rate cells
        (Table~\ref{tab:top_oscillation}) are almost exclusively
        RealSokoban levels.  Gemini-3-flash reaches
        $\bar{\rho}_\text{osc}=0.745$ on level~3, meaning nearly
        three-quarters of all steps immediately reverse the prior move.
        This reflects the agent repeatedly attempting to push the same
        box back and forth without a coherent plan—a natural consequence
        of the irreversibility constraint that distinguishes
        human-designed RealSokoban levels from procedurally generated
        ones.
  \item \textbf{Zelda and Escape}: lower bump rates ($0.30$--$0.48$)
        correlate with the higher win rates, supporting the interpretation
        that effective directional movement—not merely visiting many
        cells—is the decisive factor in reactive navigation games.
\end{itemize}

\begin{table}[htbp]
\centering
\caption{Top-10 (model, game-level) cells by mean oscillation rate
(zero-shot). High oscillation in RealSokoban reflects agents
ping-ponging boxes without a coherent plan.}
\label{tab:top_oscillation}
\small
\setlength{\tabcolsep}{5pt}
\begin{tabular}{llcr}
\toprule
\textbf{Model} & \textbf{Game-Level}
  & \textbf{Osc.\ Rate} & $n$ \\
\midrule
Gemini-3-flash  & RealSokoban-lvl3 & 0.745 & 5 \\
Gemini-3-flash  & Sokoban-lvl0     & 0.646 & 5 \\
Gemini-2.5-pro  & RealSokoban-lvl4 & 0.608 & 5 \\
o3-mini         & RealSokoban-lvl4 & 0.587 & 5 \\
o3-mini         & RealSokoban-lvl1 & 0.580 & 5 \\
Gemini-3-flash  & RealSokoban-lvl4 & 0.562 & 5 \\
o3-mini         & RealSokoban-lvl2 & 0.552 & 5 \\
DeepSeek-Chat   & Escape-lvl0      & 0.533 & 5 \\
o3-mini         & RealSokoban-lvl3 & 0.525 & 5 \\
Gemini-3-flash  & RealSokoban-lvl1 & 0.515 & 5 \\
\bottomrule
\end{tabular}
\end{table}

\begin{figure}[htbp]
  \centering
  \includegraphics[width=\textwidth]{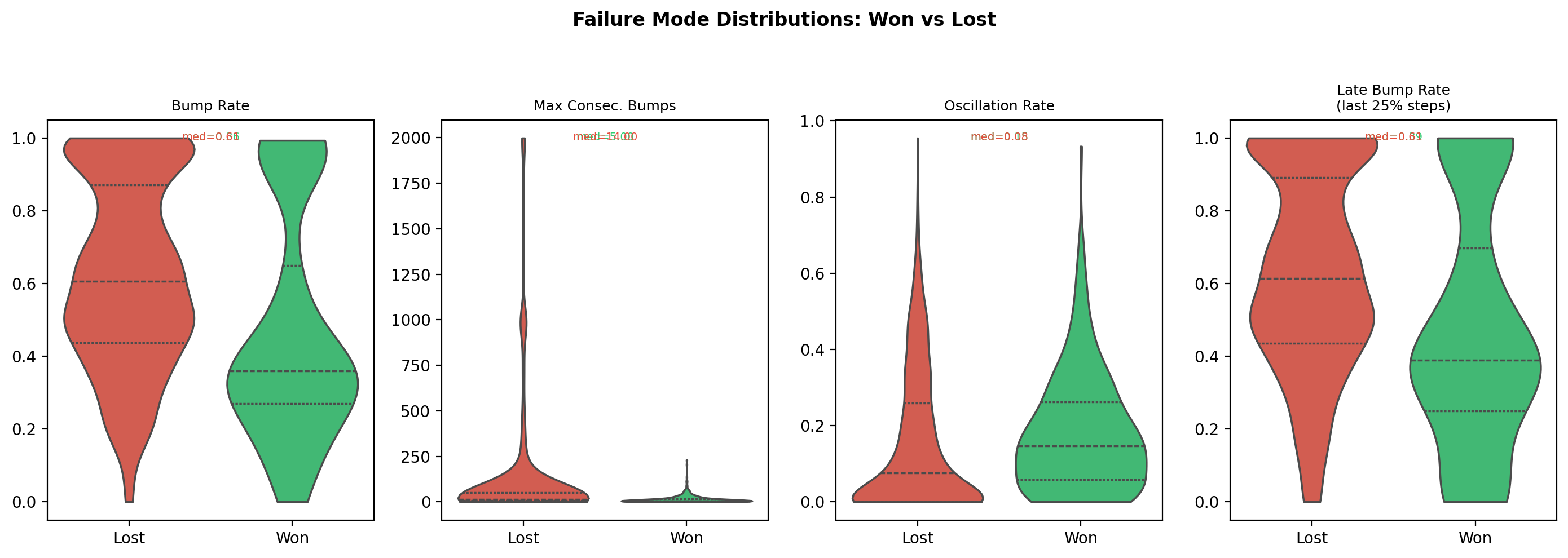}
  \caption{Violin plots of four failure-mode metrics split by episode
  outcome (green = Won, red = Lost). Bump rate, max.\ consecutive
  bumps, and late bump rate clearly separate the two outcome
  distributions; oscillation rate does not (Table~\ref{tab:failure_modes}).
  Inner lines show quartiles.}
  \label{fig:violin}
\end{figure}

\newpage
\clearpage

\newpage

\end{document}